\documentclass[10pt,twocolumn,letterpaper]{article}

\usepackage[pagenumbers]{cvpr} 

\usepackage{algorithm}
\usepackage{multicol}
\usepackage{tcolorbox}
\usepackage{siunitx}  

\newcommand{\name}{{\tt Hawkeye}\xspace}
\newcommand{\dataset}{{\tt BSTRB}\xspace}
\newcommand{\model}{{\tt Hawkeye}\xspace}

\newtcolorbox{SnackBox}[1]{
    colback=gray!10!white,    
    colframe=gray!60!black,   
    arc=8pt,                  
    boxrule=1.2pt,
    title=#1,
    fonttitle=\bfseries\small, 
    fontupper=\scriptsize\itshape,   
    coltitle=white,
    center title,
    boxsep=2pt,               
    left=4pt,
    right=4pt
}

\newcommand{\nVidCollected}{1{,}699}      
\newcommand{\nPhotoHi}{944}               
\newcommand{\nPhotoLo}{511}               
\newcommand{\nIntervals}{6}               
\newcommand{\nFrameTrain}{6{,}734}        
\newcommand{\nFrameVal}{333}              
\newcommand{\nFrameTest}{1{,}331}         
\newcommand{\nFrameSplit}{8{,}398}        
\newcommand{\nPhotoTrain}{700}            
\newcommand{\nPhotoVal}{36}               
\newcommand{\nPhotoTest}{241}             
\newcommand{\nPhotoSplit}{977}            
\newcommand{\nVidTrain}{1{,}133}          
\newcommand{\nVidVal}{56}                 
\newcommand{\nVidTest}{224}               
\newcommand{\nVidKept}{1{,}413}           
\newcommand{\nVidObjLabel}{1{,}449}       
\newcommand{\nVidSceneLabel}{1{,}559}     

\definecolor{addedcolor}{rgb}{0.00,0.40,0.18}

\renewcommand{\topfraction}{0.85}

\renewcommand{\textfraction}{0.15}
\renewcommand{\floatpagefraction}{0.70}
\renewcommand{\dbltopfraction}{0.85}
\renewcommand{\dblfloatpagefraction}{0.70}

\definecolor{cvprblue}{rgb}{0.21,0.49,0.74}
\usepackage[pagebackref,breaklinks,colorlinks,allcolors=cvprblue]{hyperref}

\usepackage[accsupp]{axessibility}  
\usepackage{orcidlink}              

\def\paperID{*****} 
\def\confName{CVPR}
\def\confYear{2026}

\title{Hawkeye: Reading Illegible Text with Vision Language Models}

\author{Brian Tang\orcidlink{0000-0001-6140-922X}\\
University of Michigan\\
Ann Arbor, MI 48105, USA\\
{\tt\small bjaytang@umich.edu}
\and
Kang G. Shin\orcidlink{0000-0003-0086-8777}\\
University of Michigan\\
Ann Arbor, MI 48105, USA\\
{\tt\small kgshin@umich.edu}
}

\begin{document}
\maketitle
\begin{abstract}
While modern Vision-Language Models (VLMs) have been remarkably 
successful in general scene text recognition, they consistently 
struggle with the long-tail distribution of text that is degraded
by motion blur or low resolution. 
These limitations are particularly apparent in egocentric applications 
like smart glasses, where text is often captured in noisy environments. 
In this paper, we introduce the \textbf{\dataset}
(Blurry Scene Text Recognition Benchmark), a novel benchmark 
of \nFrameSplit{} noisy image samples with high-fidelity annotations derived from \nVidCollected{} real-world video sequences. 
To support this, we present \textbf{\model}, a VLM fine-tuned on \dataset
to read and reason about illegible text by leveraging contextual 
priors and pixel-level alignment.
Our results show that \name matches much larger VLMs
on \dataset overall and exceeds them on text at or below 20 pixels, in a single pass
suited to real-time deployment.
We characterize the resolution at which current VLMs begin to fail.
\end{abstract}
\section{Introduction}

The advent of large language models (LLMs) has given rise to a 
suite of models capable of reasoning on diverse contextual 
information to draw concrete and fine-grained conclusions. 
Vision-language models (VLMs) have shown an impressive capability 
to reason about textual content that appears in images in 
visual question \& answer (VQA) tasks.
VLMs have outperformed and augmented the capabilities 
of traditional scene text recognition (STR) and 
optical character recognition (OCR) models in these tasks.
However, in cases where the text is far away, too small for the 
model to reliably read, or contains motion blur or compression 
artifacts, these systems inevitably struggle to produce correct STR 
outputs. This is often the case when footage of text is captured 
in real-world settings, via smart glasses, dash cameras, and even 
smartphone cameras. 
LLMs using chain-of-thought (CoT) to output reasoning tokens is 
a promising direction, but suffers from high and unpredictable latency, 
unsuitable for meeting real-time constraints.
Almost all existing STR and textual VQA datasets place 
the text in question clearly in the focus of the image, 
or only contain images where the text is visible in high resolution. 
The ground-truth text in such cases can be easily labeled 
by annotators or OCR models. However, there exists a 
long-tail distribution of low-quality text that is 
unintelligible/incomprehensible to OCR models and
VLMs, due to lack of ground truth.

\begin{figure*}[t]
\centering
\begin{subfigure}[b]{\textwidth}
  \includegraphics[width=\textwidth]{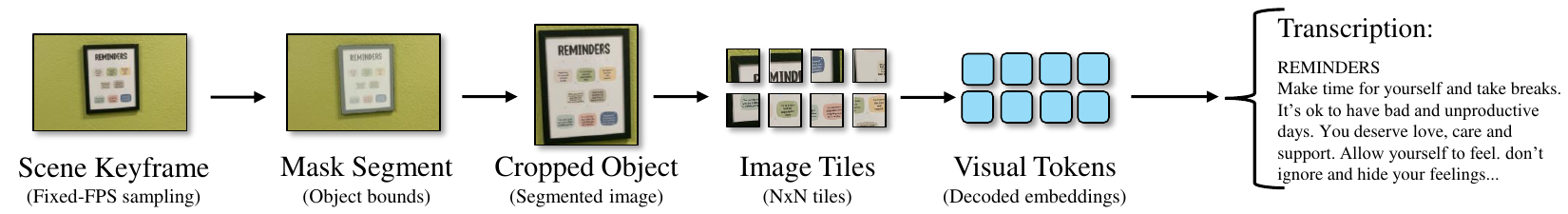}
  \caption{\name's object-centric data processing pipeline leveraging SAM3~\cite{carion2025sam} (\cref{sec:hawkeye_arch}).}
  \label{fig:architecture}
\end{subfigure}

\vspace{4pt}
\begin{subfigure}[b]{\textwidth}
  \includegraphics[width=\textwidth]{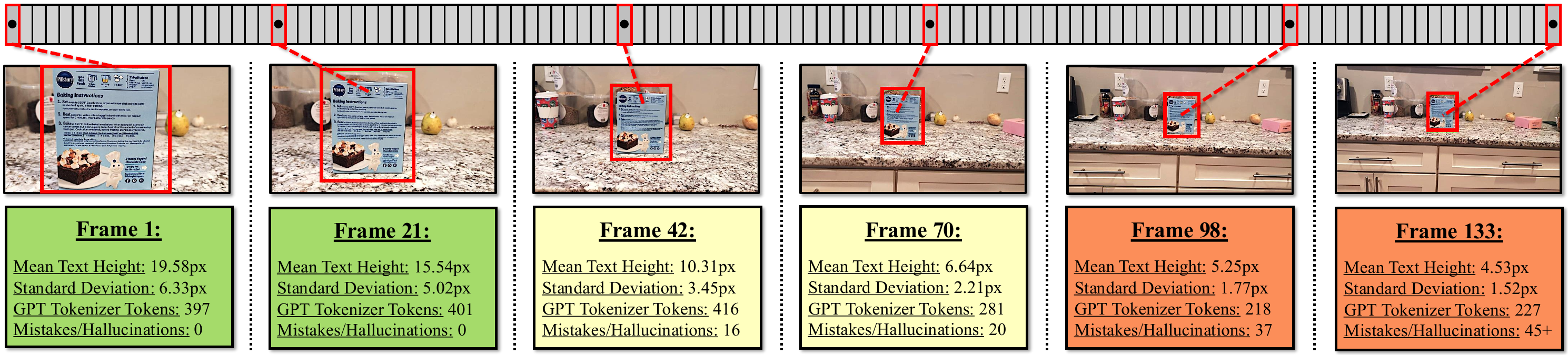}
  \caption{Frames sampled from an example \dataset video, labeled with mean text
  height, \# output tokens, and \# transcription mistakes made by GPT-5.2.}
  \label{fig:frames}
\end{subfigure}
\caption{Each video filmer walks backwards from the object, so the segmented object from the scene yields the
same text from legible to illegible (median text height 28.0\,px to
10.9\,px). Ground truth is annotated on the initial reference frames and
propagated to later frames.}
\label{fig:pipeline}
\vspace{-4pt}
\end{figure*}

Other work has largely relied on generating synthetic text-image 
datasets for training VLMs, but this has been shown to have 
significant limitations~\cite{ye2026wrongsyntheticdatascene} and can 
result in the ``model collapse'' problem~\cite{shumailov2024ai}. 
Using a novel data collection methodology, we create a new non-synthetic dataset 
that contains pairs of low-resolution images with highly accurate 
ground-truth text annotations. This dataset, the Blurry Scene 
Text Recognition Benchmark (\dataset), is built from \nVidCollected{}
recorded videos, of which \nVidKept{} survive filtering and yield
\nFrameSplit{} noisy video samples, alongside \nPhotoSplit{}
pilot photo samples (\cref{tab:counts}).
We introduce \model, a fine-tuned VLM that can read and reason 
about illegible text in images. 
It matches much larger SoTA VLMs on \dataset and outperforms
them on text at or below 20px.
\name opens up a new set of capabilities for smaller VLMs, 
enabling them to read information 
on signs, medicine bottles, flight-information boards, nutrition 
labels, papers, screens, etc., even in the presence of motion blur, 
low resolving power, occlusion, glare, and more. The VLMs can be 
deployed in real time with egocentric cameras, such as smart glasses, as
assistive tech for visually impaired users~\cite{BibEntry2026Feb,BibEntry2024Feb}.

\model is fine-tuned on \dataset video frames, first with
supervised fine-tuning and then with reinforcement learning (GRPO) that
rewards bag-of-words F1. \dataset consists of images of text from far distances, side 
angles, resulting in noisy artifacts from light diffraction,
refraction, or defocus aberration, motion blur,
and video encoding artifacts.
The ground truth images are converted into text annotations 
via SoTA OCR models and VLMs. \dataset consists of pairs of 
low-quality images of text and ground-truth text annotations 
that are used to both train \name and evaluate VLMs' 
capacity for noisy STR. 
\name therefore can learn to infer complete word sequences
nearing the acuity limit of human eyesight, achieving performance
superior to many SoTA VLMs without visual chain-of-thought.

We evaluate every model in a \emph{single-pass, non-CoT} setting: 
one prompt, one decode, no iterative zoom, crop, or re-read. The 
targeted deployment we is a wearable transcribing what its wearer is 
looking at, where a readout has to begin arriving in less than a second 
and the compute budget is a single forward pass; in contrast, the
CoT traces that lift proprietary VLMs on degraded text can take
1--7 minutes per image (\cref{sec:evaluation}). 
Our claims are, therefore, about what a model can read in one pass, 
and are not claims about the multi-pass ceilings of these models.

Reading illegible text with VLMs differs from classical 
STR/OCR super-resolution or denoising tasks in several 
fundamental ways. For example, VLMs use visual context clues 
to aid in word recognition accuracy, provide paragraph-level 
text inference, and leverage language priors (e.g., ``Nutrition
Facts'' is almost always followed by ``Calories'').
Traditional text super-resolution pipelines have been shown to
add an extra point of failure compared to the output predictions 
of a single unified vision language model. 


\name matches SoTA VLMs in text-reading accuracy at a fraction
of their size with no CoT (single-pass), suggesting that deployment 
in wearable technologies like smart glasses are practical. 
A system that robustly reads illegible text can empower the 
253 million people with visual 
impairments~\cite{bourne2017magnitude}, aid smart glasses users 
to extend the reading range, or simply reduce compute for VLM inference.

\noindent We make the following contributions with \name:
\begin{itemize}[noitemsep,leftmargin=0.4cm,topsep=5pt]
    \item \dataset, a diverse real-world training dataset and challenge 
       benchmark for blurry scene text recognition with VLMs
       (\nVidCollected{} videos, \nFrameSplit{} images; \cref{tab:counts}).
    \item A VLM using Qwen3-VL (4B and 8B) as a base, \model, fine-tuned to read even
       the smallest and noisiest of texts with improved accuracy,
       competitive with closed-weight models an order of magnitude larger, and
       ahead of them on text at or below 20px\footnote{Available at \url{https://byron123t.github.io/hawkeye/}}.
    \item A thorough evaluation comparing \name's performance
       against its base models and 20 API-served VLMs.
    \item A demonstration using \name with smart glasses,
       and the text height at which VLM reading degrades
       in the wild ($\approx$20\,px) and on clean digital text
       ($\approx$5\,px).
\end{itemize}

\noindent The following sections detail background and related 
work (\S\ref{sec:background}), \dataset's data pipeline 
(\S\ref{sec:data}), \model’s and training 
(\S\ref{sec:architecture}), and an evaluation of many VLMs' performance 
on \dataset and a demonstration 
of \name's usage (\S\ref{sec:evaluation}).

\section{Background}
\label{sec:background}

\subsection{Optical Character Recognition}

Traditional OCR has evolved from pipelines like binarization 
and layout analysis to deep learning-based systems. 
Modern OCR engines like Tesseract~\cite{smith2007overview}, 
Google Cloud Vision~\cite{BibEntry2026Febcloudvision}, and 
PaddleOCR~\cite{cui2025paddleocr30technicalreport} leverage Deep 
Neural Networks (DNNs) to extract text from primarily structured 
and semi-structured documents. However, these models often assume 
document-level clarity of text. In real-world egocentric scenery, 
where motion blur and low resolving power is prevalent, 
traditional OCR tends to perform poorly due to the lack of in-context
language and vision priors.

\subsection{Denoising and Super Resolution}

Denoising and Super-Resolution (SR) serves as an alternate approach 
to reconstructing original images from low resolution samples. 
Classical methods like SRCNN~\cite{dong2015image} and more recent 
generative approaches like Real-ESRGAN~\cite{wang2021realesrgan} 
attempt to reconstruct high-frequency details from low-quality 
inputs. There are also text super-resolution methods that focus 
on recovering character shapes to improve downstream OCR 
accuracy~\cite{wang2019textsr,zhang2024diffusion,wang2020scene}. 
However, two-stage pipelines have been reported to suffer from
error propagation, and SR models can introduce hallucinated artifacts
because they operate largely at the word level without surrounding
context~\cite{wang2020scene,zhang2024diffusion}, making them less
ideal for real-time wearable applications. 
VLMs bypass the need for an explicit reconstruction stage 
by performing inference directly on the latent features of the 
degraded image.

\subsection{Scene Text Recognition}

STR can handle word-level text recognition even in scenes where 
there exists perspective distortion, low-resolution artifacts, 
and varied fonts. Foundational works like CRNN~\cite{Jaderberg16} 
and ASTER~\cite{shi2018aster} introduced spatial transformers to 
improve distorted text recognition rates. Recent SoTA models like 
PARSeq~\cite{bautista2022parseq} and ABINet~\cite{fang2021read} 
utilize internal language models to refine visual predictions. 
Despite these advances, STR benchmarks (e.g., 
IIIT5K~\cite{MishraBMVC12}) typically contain samples of cropped 
text where the individual words or characters are the primary 
prediction outputs. \model differs by addressing the long-tail 
of blurry, sub-pixel text where the signal to noise ratio is 
insufficient for recognition without using the surrounding scene 
context provided by a VLM's multi-modal architecture.

\subsection{Vision Language Models}
\label{sec:vlm_background}

The landscape of VLMs has shifted from simple vision 
encoders~\cite{radford2021learning} to unified generative 
architectures. Early work in visual instruction tuning, such as 
LLaVA~\cite{liu2023visual}, demonstrated that visual features 
could be aligned to LLMs as word embeddings. 
VILA~\cite{lin2024vila}, NVILA~\cite{liu2025nvila}, 
Qwen2.5-VL~\cite{bai2025qwen2}, and Qwen3-VL~\cite{bai2025qwen3} 
have demonstrated that VLM performance has been scaling quickly. 
Many of these models also support variable-resolution inputs, 
supposedly allowing models to process fine-grained textual details 
that were lost in fixed-dimension resizing. Subsequent iterations 
have focused on scaling both resolution and efficiency. VILA-HD 
scaled vision pre-training to 4K resolution~\cite{shi2025scaling}.

\subsubsection{VLM Architectures.}

Modern VLMs couple a \emph{vision encoder} with a \emph{language 
model} to generate or understand text grounded in images. 
An image $I$ is split into patches and encoded by a ViT-style 
encoder $E_{\theta_v}$ to produce visual embeddings. 
A lightweight \emph{projector} $P_{\theta_p}$ maps these 
embeddings into the language model’s token space, yielding 
a set of visual tokens (keys/values) $K$ for cross-attention:

\[
K \;=\; P_{\theta_p}\!\big(E_{\theta_v}(\mathrm{patchify}(I))\big).
\]

A causal decoder $D_{\theta_\ell}$ then conditions on $K$ and an 
optional text prompt $p$ to produce the output sequence $y$ in an 
autoregressive manner. Training is typically next-token prediction 
on image-text pairs with instruction tuning. At inference, the 
model decodes $\hat y = \arg\max_y P_\theta(y \mid I, p)$ using 
beam search or greedy decoding. Cross-attention lets each 
generated token reference \emph{all} image-derived tokens $K$ and 
the full text prefix. \name\ follows this standard recipe (vision 
encoder $\rightarrow$ projector $\rightarrow$ language model), fine-tuning the vision encoder, projector, and language model, then
refining it with reinforcement learning.

\section{System Design}
\label{sec:design}

\begin{figure*}[t]
\centering
\begin{subfigure}[b]{0.55\textwidth}
  \includegraphics[width=\textwidth]{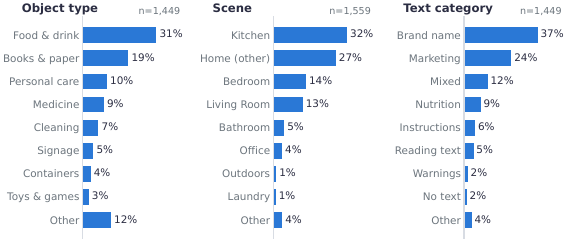}
  \caption{Object, scene, and text category distributions.}
  \label{fig:dataset_composition}
\end{subfigure}\hfill
\begin{subfigure}[b]{0.43\textwidth}
  \includegraphics[width=\textwidth]{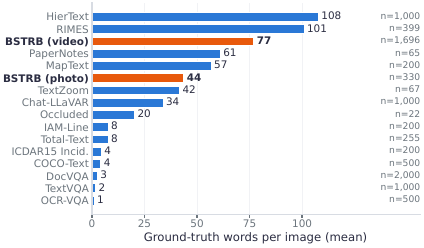}
  \caption{Ground-truth words per image; \dataset in orange.}
  \label{fig:text_density}
\end{subfigure}
\caption{\dataset composition and text density. Panel (a) is over the clips
carrying each label (\nVidObjLabel{} object/text, \nVidSceneLabel{} scene) of
\nVidCollected{} collected; see \cref{tab:counts}. \dataset is denser than
every benchmark except HierText and RIMES.}
\label{fig:dataset}
\vspace{-4pt}
\end{figure*}


\subsection{Problem Formulation}
\label{sec:hawkeye_formulation}
Let $I$ denote a scene image containing text like signboards, 
labels, screens, etc., and $y=\{y_1,\ldots,y_N\}$ the 
ground-truth text transcription(s) for a sequence of words or 
tokens that appear in the scene. $P_\theta$ is a probability 
distribution derived from a DNN, and $\theta$ denotes the 
trainable parameters of our model. Standard STR and OCR models 
aim to maximize $P_\theta(y \!\mid\! I)$ where
\begin{equation}
P_\theta(y \mid I) ~\approx~ \sum_{b} P_\theta(b 
\mid I)\;\prod_{i=1}^N P_\theta(w_i \mid I, b_i).
\label{eq:ocr_factor}
\end{equation}
Conventional STR/OCR pipelines assume conditional independence 
across words, decoding each region in isolation.
Context, when used, is typically injected post-hoc by passing 
the raw OCR outputs into a language model to fix spelling mistakes 
and enforce grammatical correctness (without image grounding).

Conversely, VLMs are often trained on images, primarily with 
documents and text to improve their OCR capabilities. 
These VLMs attain generalizable text-reading capabilities, 
but their capabilities are much more limited ``in the wild'' 
than the performance they achieve with document parsing.

\subsection{System Architecture}
\label{sec:hawkeye_arch}
We condier a system that is comprised of a VLM extended with an 
object proposal and segmentation stage (Fig.~\ref{fig:architecture}):

\begin{enumerate}[noitemsep,leftmargin=0.5cm]

\item \textbf{Object Proposer.} We employ a first pass to a 
smaller VLM to identify candidate object labels $R$ in the scene.

\item \textbf{Segmentation Module (SAM3).} We utilize 
SAM3~\cite{carion2025sam} (Segment Anything Model 3) to 
refine proposals $R$ into high-fidelity masks $MR$. 
This isolates text and reduces information 
loss from the VLM rescaling the image.

\item \textbf{Vision Encoder.} A ViT encoder maps 
masked image regions $I\odot MR$ to a sequence of visual 
tokens $v_{1..M}$.

\item \textbf{Projector.} A projection layer stack aligns 
visual tokens to the LLM token space.

\item \textbf{LLM Backbone.} An autoregressive decoder consumes 
$v_{1..M}$ and produces the transcription tokens.
 
\end{enumerate}

\noindent Stages 3--5 operate on the crop $I \odot MR$: \model reads a
segmented region at both training and test time.

\section{\dataset Data Collection and Annotation}
\label{sec:data}

\subsection{Data Collection Study}

We conducted a study to collect diverse 
data of text-bearing objects across varying resolutions, 
camera angles, etc. Our study was approved and exempted from 
oversight by our Institutional Review Board (IRB). 
We recruited participants through Prolific, with the following 
requirements: 18+ years old, fluent in English, located in the U.S., 
use a mobile device with a camera\footnote{Recruitment criteria, 
consent form, instructions, etc., are in \cref{sec:appendix_study}.}.

Our final approved sample consisted of 244 participants.
The median completion time for the study was 12:05 minutes, 
with an average compensation rate of \$9.93/hr. 
Our sample was skewed towards middle-aged participants (median=37.68 
years, range: 18--76 years, SD=10.54) and was relatively balanced 
by gender (111 men, 130 women, 2 other). Participants had a 
diversity of ethnicity, education levels, and employment statuses. 
The entire dataset collection study, including the pilot-tests, 
cost \$818.34 to run.

\subsection{Image and Video Collection}

Participants were instructed to complete
tasks designed to capture images of textual content at 
varying resolutions:
\begin{enumerate}
\item \textbf{Photo Pairs via Mobile Phone:} 
Participants took two static photos
of an object containing text. The first was a ``close-up'' to 
capture a high-resolution image of the ground truth text, and 
one was taken ``farther away'' to simulate real-life resolution 
and environment conditions.

\item \textbf{Video Recordings via Mobile Phone:} Participants
began with a clear, close-up shot of the text-bearing object and
recorded continuously while walking backward for roughly 8 seconds. 
This allowed us to track the object through the frames, to pair 
the lower resolution and lower scale image frames of the object with 
the initial ground-truth shot of the object and its text.
\end{enumerate}

They were given a demonstration photos and videos. 
The participants were instructed to film contents containing text 
in a similar fashion, provided that no personal identifying 
information was present in the scenes (e.g., mail, address, bills, 
prescription pills, faces, etc.). 
Finally, they were instructed to take footage of objects 
containing text with 10 or more words. Participants were 
instructed to take either 6 videos or 15 photos of text. 
We received a total of \nVidCollected{} videos, 
\nPhotoHi{} high-res photos, and \nPhotoLo{} low-res photos.
\cref{tab:counts} reports pre/post-filtering counts.

\subsection{Automated VLM Annotations}

To scale and automate our annotations, 
we utilized a data processing pipeline. 
For each sample, we designated the initial high-resolution 
frame (or the close-up photo) transcribed by GPT-5\footnote{gpt-5-2025-08-07~\cite{singh2025openai} (most VLMs can perform this transcription)} as the source 
of ground truth. We manually verified and corrected the ground truth of the
full held-out set (the golden set: 280 videos and 277 photos), from which the validation
and test splits are drawn (inter-annotator agreement in \cref{tab:interannotator_agreement}). 
Review changed 20\% of the video labels and 56\% of the photo labels, mostly because the
automatic photo labels were incomplete; 2 flagged videos were dropped. Training labels are
left as produced by the pipeline. GPT-5 was used to generate short 3--5 word 
phrase descriptions (e.g., ``red hot sauce bottle'' or ``book with big text'') 
to be used downstream as inputs for SAM3 to segment 
and track the text-bearing objects. 

\begin{figure*}[t]
\centering
\includegraphics[width=\textwidth]{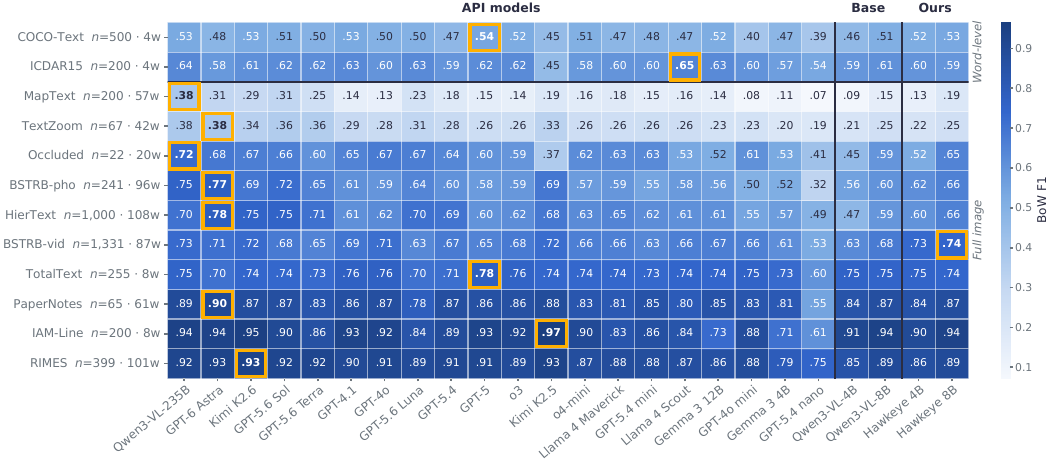}
\caption{Transcription benchmark results, BoW F1. Columns: 20 API models (by mean), Qwen3-VL bases, and Hawkeye. Rows: 10 OCR datasets and held-out \dataset, grouped by ground-truth length, with $n$ and mean ground-truth words. Boxed: best per row.
GPT-6 Astra is best on 4 of 12 columns; Hawkeye (Qwen3-VL-8B) is best on \dataset-vid.}
\label{fig:heatmap_f1}
\vspace{-4pt}
\end{figure*}

\subsection{Object Tracking and Consistency}

To maintain the association between the ground-truth text and video frames,
we used the Segment Anything Model 3 (SAM3) to track the object's 
bounds. This minimizes text uncaptured by initial frames that may misalign annotations.

To ensure tracking integrity and to filter out drift, we calculated 
the Intersection over Union (IoU) between consecutive frames 
to select the mask to continue using. This reduces issues from multiple
object masks being proposed in later frames.
Masks proposed by SAM3 that fell below the confidence score threshold of 0.3
were discarded. Our dataset processing pipeline extracted frames at the 
following \nIntervals{} intervals of the videos: 0.01, 0.15, 0.30, 0.50, 0.70,
0.95. Intervals at which SAM3 emits no mask for a tracked object are
skipped (\cref{tab:counts}).
\cref{fig:pipeline} contains a summary of our dataset pipeline and example of the sampling intervals.

\subsection{Text Annotation and Text Quality Metrics}
\label{sec:gt_propagation}

Text in later frames is too noisy for OCR to detect, so we estimate text
height instead. Google Vision OCR is used to detect word
bounding boxes on $f_0$ and obtain the median height $h_0$. For later
frames $t$, we scale $h_0$ by the change in the SAM3 object bounds:
\begin{equation}
\hat{h}_t = h_0 \cdot \frac{s_t}{s_0}, \quad s_t = \sqrt{W_t H_t},
\label{eq:height}
\end{equation}
where $W_t$ and $H_t$ are the width and height of the bounding box.
We removed samples that had no text
in $f_0$ or no SAM3 bounds (\cref{tab:counts}).
Each sample includes derived metadata on metrics like $\hat{h}_t$, 
image sharpness (Laplacian variance),
perceptual similarity to $f_0$ (LPIPS~\cite{zhang2018perceptual}), 
and SIFT feature matches to $f_0$~\cite{lowe1999object}
(\cref{sec:appendix_dataset}).

\subsection{Dataset Details}

The \dataset dataset consists of primarily food and beverage packaging 
(31\%) in kitchen contexts (32\%), likely due to the accessibility 
and high presence of text on such items. This was followed by books, 
magazines, paper (19\%) and in other home contexts (27\%). Most 
samples contained text related to product names, description and 
marketing, ingredients and nutrition, or multiple categories. 
A breakdown of the dataset is available in
\cref{fig:dataset_composition}.

\begin{table}[t]
\centering
\caption{\dataset counts at each pipeline stage. Within a surviving clip, each tracked object is sampled at \nIntervals{}
intervals (\cref{sec:data}).}
\label{tab:counts}
\footnotesize
\setlength{\tabcolsep}{4pt}
\begin{tabular}{llrr}
\toprule
\textbf{Stage} & \textbf{Unit} & \textbf{Video} & \textbf{Photo} \\
\midrule
Collected               & clips / photos & \nVidCollected & \nPhotoHi\,+\,\nPhotoLo \\
Retained after filtering & clips         & \nVidKept      & --- \\
\midrule
Train (automatic labels) & frames (videos) & \nFrameTrain{} (\nVidTrain) & \nPhotoTrain \\
Validation (verified)    & frames (videos) & \nFrameVal{} (\nVidVal)     & \nPhotoVal \\
Held-out test (verified) & frames (videos) & \nFrameTest{} (\nVidTest)   & \nPhotoTest \\
\bottomrule
\end{tabular}
\vspace{-4pt}
\end{table}

The average number of tokens and words per image sample is 144.1 and 
75.5 respectively (max 2,171 and 753). The total tokens and words 
in our dataset is 1,240,055 and 649,861, respectively. Prior noisy-text 
benchmarks (e.g., TextZoom) are lacking in either word-density or sample size.
At a mean of 77 ground-truth words per video
sample \dataset is exceeded only by HierText and RIMES (which is just a
handwriting dataset). Median text height falls from
28.0px in the reference frame to 10.9px in the last, and 50.0\% of
frames contain text
under 20px; frame-level degradation statistics are given in
\cref{sec:appendix_dataset}. Train, validation and test for \dataset (videos) are
split with no \texttt{base\_id} overlap (\nVidTrain{} / \nVidVal{} / \nVidTest{} videos,
\cref{tab:counts}).

\section{Architecture and Training}
\label{sec:architecture}

\begin{figure}[t]
\centering
\includegraphics[width=\columnwidth]{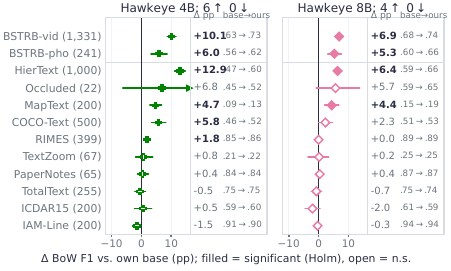}
\caption{Hawkeye 4B and 8B $-$ their Qwen3-VL bases: paired differences in BoW F1 with 95\%
CIs, clustered by video~\cite{miller2024adding}; filled: significant after Holm. Right of each
panel: the difference (bold if significant) and base $\rightarrow$ Hawkeye BoW F1.}
\label{fig:forest_base}
\vspace{-4pt}
\end{figure}

\subsection{Training Data and Augmentation}

\model is trained on \dataset video frames: \nFrameTrain{} frames from 
\nVidTrain{} videos at the six sampled positions, labeled by the automatic 
pipeline of \cref{sec:data}, each with its own \dataset instruction. 
We also tried training with the \dataset photos and the public OCR and QA sets 
of \cref{tab:dataset_breakdown}, on the validation splits, but this ended up being a detriment to overall performance.
We split \dataset by video id into train, validation (\nVidVal{} videos) and 
test (\nVidTest{} videos) with no overlap (\cref{tab:counts}).
The validation split was used to select the training recipe.

Some of the datasets we use for evaluation have overlap with 
the OCRBench dataset~\cite{Liu_2024,fu2024ocrbenchv2improvedbenchmark}, 
but we modify input and output tokens to focus more on text transcription 
compared to general text VQA and OCR bounding box detection. Unlike OCRBench, 
which primarily tests recognition and reasoning on easily visible text, 
and unlike TextZoom~\cite{wang2020scene}, which tests word-level 
reconstruction in the wild,
\dataset requires models to perform reasoned reconstruction of both long 
textual passages and visual logos and patterns.
For instance, in a blurred image of a medicine bottle, \model needs its 
linguistic backbone to identify ``Dosage'' even if the 'D' and 'g' are 
visually indistinguishable. For models to succeed on \dataset, they will
need to succeed at drawing correlations between text regions.

\subsection{Architecture}

We fine-tune Qwen3-VL~\cite{bai2025qwen3} at 4B and 8B parameters. Both accept variable-resolution
input. It fuses ViT features through DeepStack integration, uses interleaved-MRoPE for
spatial-temporal modeling and 16-pixel patches, and jointly pre-trains text and visual
modalities from an early stage. We train both sizes with the same recipe so that every
result can be checked at two language model scales.

\subsection{Training Procedure}

Training has two stages (\cref{sec:appendix_training_data}). First, supervised
fine-tuning updates the whole model with AdamW, using learning rates of $10^{-6}$
(language model), $10^{-5}$ (projector) and $10^{-6}$ (vision encoder), a cosine schedule
with 3\% warmup, an effective batch size of 32, and 420 steps ($\approx$2 epochs); we keep
the checkpoint with the lowest validation loss. Second, we apply
GRPO~\cite{shao2024deepseekmath,liu2025understanding} with a fresh rank-64
LoRA~\cite{hu2022lora} on the language model's attention projections, sampling 8
completions per frame and using each sample's $F_{bow}$ (\cref{sec:metrics}) as the
reward, with a KL penalty of 0.02 against the SFT model, for 300 steps. Every image is
upscaled $\times$3 (bicubic) before the processor's pixel budget of 65,536--1,003,520, at
both training and test time. On validation, SFT followed by GRPO gains $+8.2$pp video BoW
F1 over a preliminary version of \name (Qwen3-VL-4B fully fine-tuned on a mixture of 13 OCR
datasets and \dataset), against $+3.2$pp for SFT alone and $+5.1$pp for GRPO alone.

\noindent\textbf{Hardware Setup.} 
Training, evaluation, and deployment is performed on a private server cluster equipped with 
$12\times$ NVIDIA A6000 Ada GPUs (48GB VRAM each).
\section{Evaluation}
\label{sec:evaluation}

We evaluate \model against the baseline open-source model it's trained on and a suite of
proprietary VLMs to assess its proficiency in general
OCR and \dataset-related tasks.
All 24 models (20 API, 4 local) receive the identical cropped images and system prompts.

\subsection{Evaluation Metrics}
\label{sec:metrics}

We evaluate performance using 3 metrics calculated on lowercased, 
punctuation-stripped token sequences for ground truth $g$
(length $N$) and hypothesis $h$ (length $M$).

\paragraph{Positional Metrics (WER \& Aligned F1).}
We measure structural accuracy via Levenshtein distance. 
Let $S$,$I$,$D$ be substitutions, insertions, and deletions, 
and $C=N-S-D$ be the number of correct matches. We report:
\small
\begin{equation}
\mathrm{WER} = \frac{S + I + D}{N}, \quad P_{\mathrm{align}} = 
\frac{C}{M}, \quad R_{\mathrm{align}} = \frac{C}{N}.
\end{equation}
The aligned F1 ($F_{align}$) is the harmonic mean of $P_{align}$ and
$R_{align}$, penalizing both hallucinations and misorderings. One issue with positional metrics, is that there is often not a well-defined reading order for objects with multiple text regions.

\paragraph{Lexical Metrics (WIR \& BoW-F1).}
To isolate content recall from spatial/ordering errors, we use 
multiset intersection $O=\sum w\cdot min(C_g \cdot (w), C_h \cdot (w))$ where $C(w)$
is the word frequency:
\small
\begin{equation}
\mathrm{WIR} = \frac{O}{N}, \quad P_{\mathrm{bow}} = \frac{O}{M}, 
\quad R_{\mathrm{bow}} = \frac{O}{N}.
\end{equation}
Word Inclusion Rate (WIR) serves as a lenient recall measure, while 
the Bag-of-Words F1 ($F_{bow}$) captures lexical overlap independently 
of position. The delta between $F_{bow}$ and $F_{align}$ specifically 
highlights errors attributable to word-order mistakes versus 
recognition failures. $F_{bow}$ penalizes verbosity and hallucination
through precision, whereas WIR does not, so we report both.

\subsection{Evaluation Protocol}
\label{sec:protocol}

We report a \textit{transcription track} of 12 datasets: 10 general OCR sets and \dataset
video and photo. Every model sees the same images, so we report per-sample paired differences
with standard errors clustered by source video (\dataset) or chart
(Snellen)~\cite{miller2024adding}. All intervals are 95\%
bootstrap CIs and all $p$-values are Holm-corrected. Outputs are scored raw; stripping
markdown, preambles and bullets first moves no model's mean by more than 0.07pp
(\cref{sec:appendix_protocol}).

\subsection{OCR and STR Benchmarks}

We first establish \model's performance on standard text benchmarks to
ensure that fine-tuning for blurry text does not lead to catastrophic
forgetting of general OCR capabilities
(\cref{fig:heatmap_f1,fig:heatmap_wir}).
No single model leads the transcription track: GPT-6 Astra is best on 4 of 12 datasets under
$F_{bow}$, and \name 8B is best only on \dataset-video. Against its own base model (\cref{fig:forest_base}), \name 4B improves significantly on
6 of 12 datasets and \name 8B on 4 datasets, with no significant regression at either size, even
though \name is trained on \dataset alone. Outside
\dataset the gains are on HierText ($+12.9$pp at 4B, $+6.4$pp at 8B), MapText and, for 4B,
COCO-Text and RIMES; the remaining OCR sets are within a few points of the base
(\cref{fig:forgetting}).

\subsection{Blurry Scene Text Recognition Benchmark}
\label{sec:bstrb_bench}

\begin{figure}[t]
\centering
\includegraphics[width=\columnwidth]{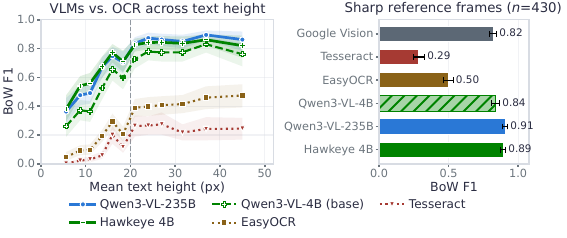}
\caption{Dedicated OCR engines compared to VLMs on \dataset. Google Vision is competitive
only on the sharp reference frames.}
\label{fig:ocr_vs_vlm}
\vspace{-4pt}
\end{figure}

\noindent\textbf{OCR Engine baseline.} On sharp reference frames
traditional OCR is competitive with VLMs (Google Vision $.822$ BoW F1 against
$.892$ for \name 4B). However, on all frames, performance collapses to
$.296$ (EasyOCR) and $.162$ (Tesseract), against $.728$ for \name 4B (\cref{fig:ocr_vs_vlm}).

\begin{figure*}[t]
\centering
\begin{subfigure}[b]{0.39\textwidth}
  \includegraphics[width=\textwidth]{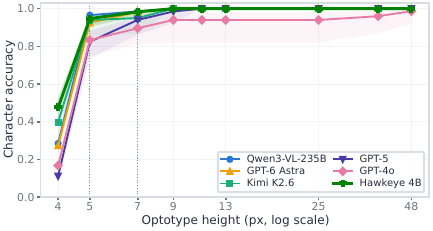}
  \caption{SoTA VLM performance on Snellen}
  \label{fig:snellen_api}
\end{subfigure}\hfill
\begin{subfigure}[b]{0.39\textwidth}
  \includegraphics[width=\textwidth]{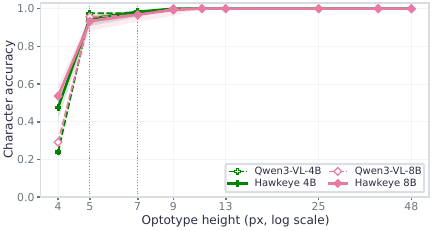}
  \caption{\name performance on Snellen}
  \label{fig:snellen_os}
\end{subfigure}\hfill
\begin{subfigure}[b]{0.2\textwidth}
  \includegraphics[width=\textwidth]{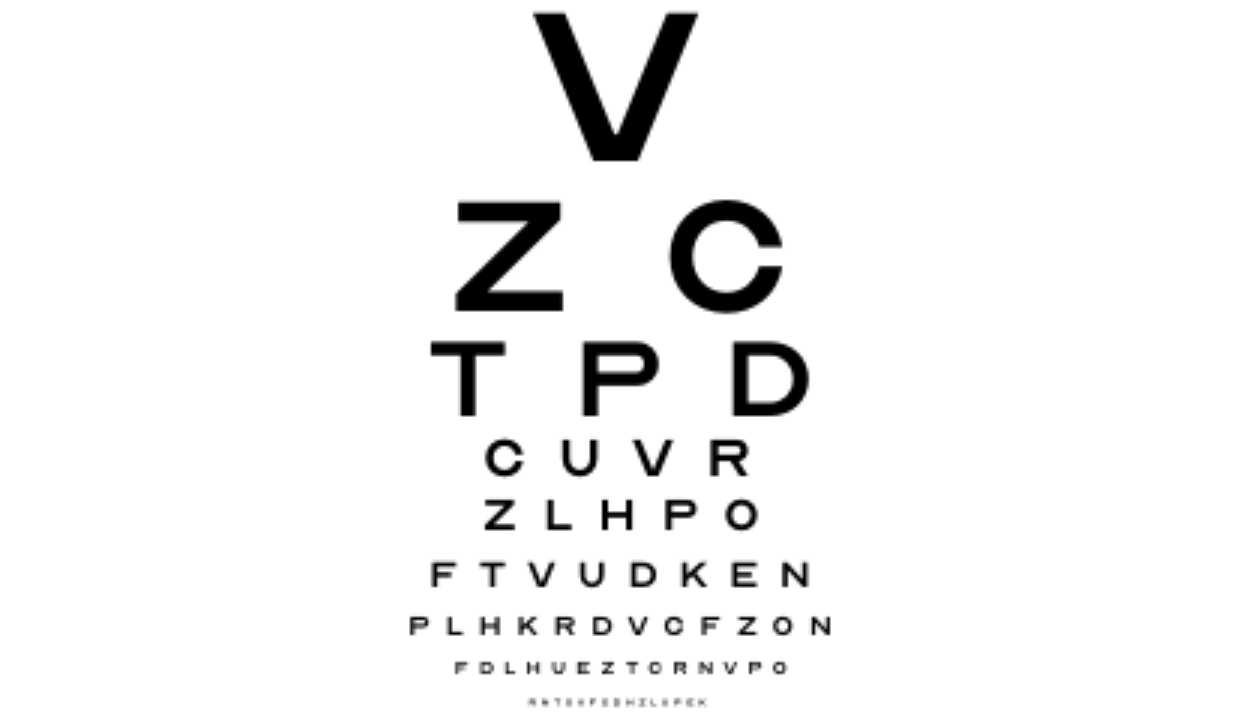}
  \caption{(1/50) Example Snellen chart used}
  \label{fig:snellen_ex}
\end{subfigure}\hfill
\caption{SoTA proprietary VLMs for OCR on Snellen charts. 
Performance drops at 5px.}
\label{fig:snellen}
\end{figure*}

\begin{figure*}[t]
\centering
\includegraphics[width=\textwidth]{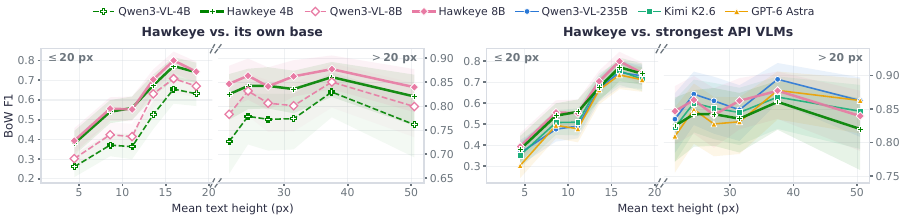}
\caption{BoW F1 on held-out \dataset video frames (1,331 frames,
224 videos) by mean text height, split at 20\,px with a separate y-scale on each
side. Left: Hawkeye (solid) vs.\ its base (dashed, same colour). Right: Hawkeye vs.\ the
strongest API VLMs. Equal-count bins ($n=98$--$111$), bootstrap 95\% CI bands.}
\label{fig:bstrb}\label{fig:bstrb_f1}\label{fig:bstrb_wir}
\vspace{-4pt}
\end{figure*}

\noindent\textbf{Text Size.} For ``in-the-wild'' images (e.g.,
\dataset), VLM performance begins to degrade at around 20 pixels of mean text
height (\cref{fig:bstrb}); on clean, generated digital text the same models
hold accuracy down to roughly 5 pixels, $4\times$ smaller
(\cref{fig:snellen}). \name improves performance at this
boundary (\cref{fig:bstrb}). For frames with mean text height at or below
20px, \name 4B gains $+14.2$pp BoW F1 and \name 8B $+9.9$pp over their bases.

\noindent\textbf{Clean digital text.} As a control, we render 50 pages of generated Snellen
charts~\cite{snellen1873probebuchstaben} and prompt each model with the item's prompt only
(\cref{fig:snellen}). Most API models and every Qwen3-VL and \name model keep at least 90\%
character accuracy down to about 5\,px; weaker or older models stop between 6 and 8\,px
(GPT-4.1 and Llama~4 6.0, GPT-5 6.3, o3 6.6, GPT-4o 7.2\,px). GPT-4o mini and Gemma~3 4B write
one character per line, which breaks the line alignment of the scorer, so their curves reflect
formatting rather than reading. All models are near ceiling from 7\,px, so differences live at
4--5\,px: at 4\,px \name 4B reaches .48 character accuracy against .40 for Kimi K2.6, .28 for
Qwen3-VL-235B, .27 for GPT-6 Astra and .11 for GPT-5. Against their own bases, clustered by
chart, both \name sizes gain on the smallest optotypes: $+7.2$pp (4B, $[+4.5, +9.9]$) and
$+7.5$pp (8B, $[+4.7, +10.2]$) at or below 7\,px, both $p_{\mathrm{Holm}}<.001$, and $+2.4$pp
over all rows at both sizes.

\cref{fig:bstrb} also indicates the best performing API VLMs for STR are Qwen3-VL-235B,
GPT-6 Astra, and Kimi K2.6. Over the full test split (1,572 items), \name 8B is roughly tied
with them ($.730$ BoW F1 against $.731$, $.719$ and $.718$; n.s.). On video frames at or below
20px it beats all three significantly ($+3.7$ to $+6.0$pp). On the pilot photos, which are
excluded from training, it is behind ($.655$ against $.773$ for GPT-6 Astra).

\noindent\textbf{Reading vs.\ guessing.} \cref{fig:error_taxonomy} assigns each ground-truth
word to correct, misread, missed, or inserted. \name tends to convert misses into correct reads
while also emitting less text: correct words rise from 57\% to 70\% (4B) and from 61\% to 70\% (8B),
and inserted words per ground-truth word fall from 1.83 to 0.80 and from 1.26 to 0.60. Not just that, but \name also tends to fall into fewer repetition loops (\cref{sec:appendix_errors}).

\section{Discussion and Ethical Concerns}
\label{sec:discussion}

\subsection{Out-of-Distribution Generalization}
\label{sec:ood}

\dataset is collected from everyday household objects and \model both
trains and infers on single-object crops. One question that arises, is
whether it has overfit to that distribution, rather than generalizing to learn to read degraded text. We tested \name 8B and its base on 37 objects (91 images) captured
with smart glasses mostly outside the training distribution. This includes screen text,
outdoor signage, printed documents, a lab environment, and Chinese characters (\cref{fig:ood}). The result is still positive: word inclusion improves by $+13.0$pp
($[+6.9, +19.7]$), BoW F1 by $+11.5$pp and aligned F1 by $+12.3$pp (all
$p_{\mathrm{Holm}}<0.001$), and 26 of 37 objects improve against 5 that regress
(\cref{fig:ood_by_folder}).

\subsection{Attention Probes on Hawkeye}
\label{sec:blind_spots}

To understand \emph{why} \model reads degraded images more accurately, we measure attention to
image patches for each generated token (last four layers, attention sinks excluded) on 160
held-out frames, 141 of them with automatically detected text. \model's processor is pinned
to 1.0\,MP ($\approx$1.2k visual tokens) against a ceiling of 16.8\,MP for Qwen3-VL, after a
$\times$3 upscale; both arms here see the same input. \model does not cover more of the text
than its base: it visits as many detected text regions (base $\rightarrow$ \model: 4B
$.85 \rightarrow .84$, 8B $.80 \rightarrow .80$), and covering more regions does not predict
accuracy. What changes is where the extra reading occurs. On text at or below 20px its
attention shifts from the large print toward fine print that the base Qwen models skip ($-6.5$pp
of attention on the detected large print at 4B, $-5.7$pp at 8B; about $-1$pp above 20px),
and frames with the larger shift are those where it transcribes more words
(\cref{fig:showcase}). Fine-tuning also moves attention off content-free sink patches
($-3.2$pp at 4B, $-4.2$pp at 8B). \model thus is able to read small text, where
base models settle on the largest text (\cref{sec:appendix_attn}).

\subsection{Limitations and Future Work}

\noindent\textbf{No super-resolution baseline.} 
We motivate a single-stage model partly by the behavior reported for
SR$\rightarrow$OCR pipelines, but do not test or run one on \dataset due to the scope of this work.

\noindent\textbf{Transcriptions have no markup.} \dataset and \model do not record any styling markup. Since we are targeting an edge device producing a spoken or plain-text readout, omitting markup both keeps annotations more consistent and lowers token usage.

\noindent\textbf{Single-pass only.} We report no visual
chain-of-thought or agentic configuration for models. 
This keeps the comparison scoped to the real-time budget for our use case. This means our results are on the lower
bound on what the proprietary models can do given unbounded test-time. 
Performance substantially improves when used with 
visual chain-of-thought, but costs 1--7 minutes per image.

\noindent\textbf{Unverified training labels.} Training labels are automatic transcripts;
only the validation and test labels are human-verified. Because labels are per-video, a
degraded frame's label could include text the frame no longer shows which would reward guessing.

\noindent\textbf{Future work directions.} Future work can focus on expanding the dataset, extending its coverage to multilingual support, including user studies with visually impaired 
participants to assess the trade-offs between \name and commercial VLMs.

\subsection{Responsible AI and Dataset Use}

\noindent\textbf{Privacy and Responsible AI.}
Data collection carries inherent de-anonymization risks if video
samples are aggregated. Participants were informed of these risks, 
and PII was sanitized. These risks are outweighed by the potential 
benefits of this research to create new assistive technologies. 
We acknowledge the potential for malicious use cases and 
emphasize responsible deployment.

\noindent\textbf{Dataset \& Model Release.}
To help mitigate downstream harms, \dataset and \name will be
publicly released via gated access. Researchers must provide 
institutional emails for verification and sign
a Data Use Agreement prohibiting de-anonymization.
\section{Conclusion}
\label{sec:conclusion}

We have presented a limitation of current VLMs: reading 
distant, blurry, and degraded text in practical egocentric camera settings. 
We 
introduce \dataset, a novel real-world dataset and challenge benchmark specifically
curated for blurry scene text recognition.
We train
\model, a VLM fine-tuned to synergize pixel-level visual features 
with contextual language priors for scene text 
transcription. \name demonstrates quantitative and qualitative improvements in tasks related to STR and OCR, and matches the performance of the strongest VLMs on the market.

{
    \small
    \bibliographystyle{ieeenat_fullname}
    \bibliography{main}
}

\clearpage
\setcounter{page}{1}
\maketitlesupplementary

\setcounter{totalnumber}{5}
\setcounter{topnumber}{3}
\setcounter{dbltopnumber}{3}
\renewcommand{\floatpagefraction}{0.6}
\renewcommand{\dblfloatpagefraction}{0.7}
\renewcommand{\topfraction}{0.9}
\renewcommand{\dbltopfraction}{0.9}
\renewcommand{\textfraction}{0.07}

\section{Study Protocol and Materials}
\label{sec:appendix_study}

\begin{figure*}[p]
\centering
\includegraphics[width=\textwidth]{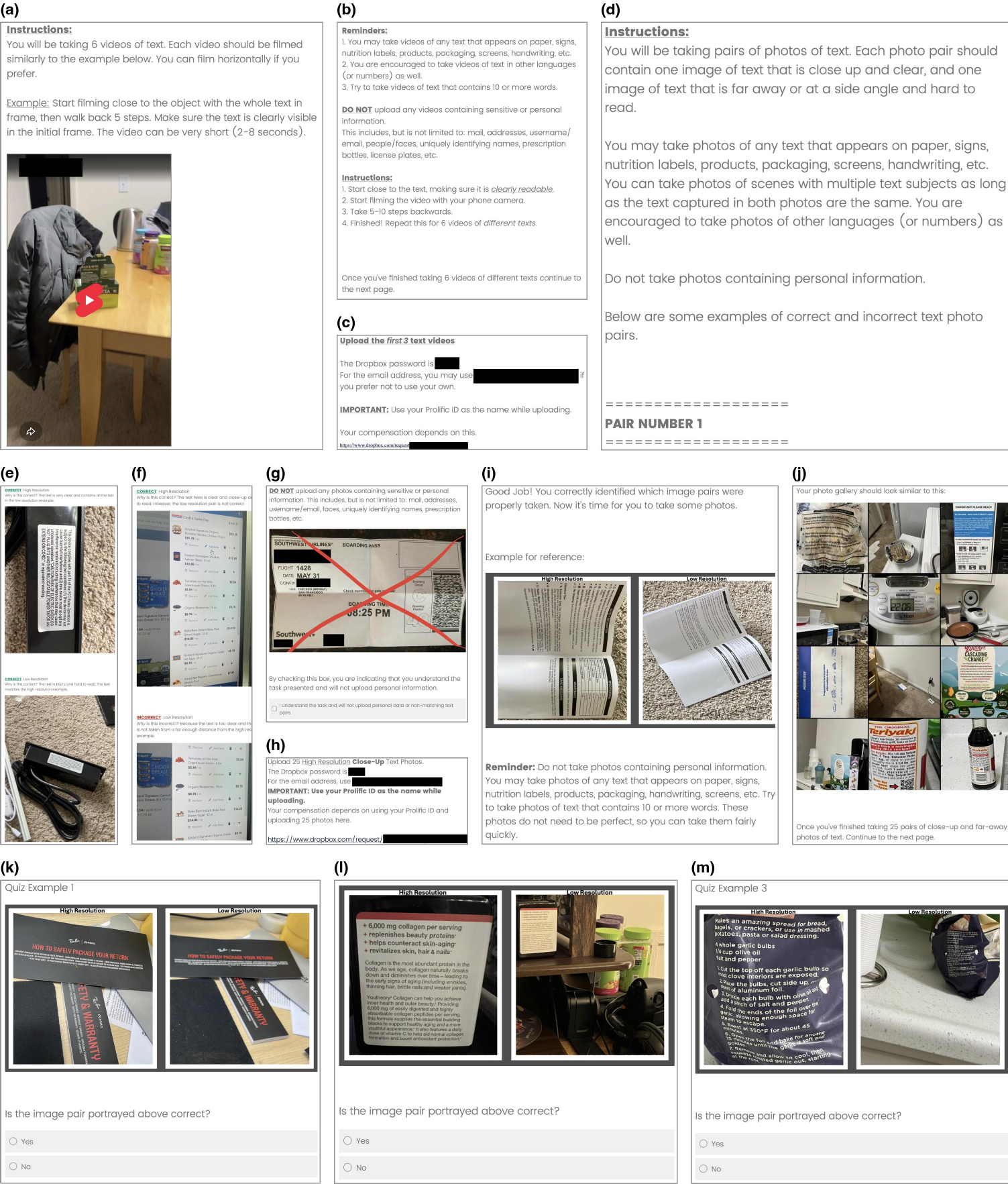}
\caption{Study pages shown to Prolific participants, in survey order; identifying details
(study account, e-mail address, upload links) are redacted. \textbf{Video study (\dataset):}
(a)~instructions with the demonstration video, (b)~reminders and steps, (c)~upload page.
\textbf{Photo pilot:} (d)~instructions; (e)~a correct and (f)~an incorrect example pair, each
with its explanation; (g)~privacy acknowledgement; (h)~upload page; (i)~reference pair and
reminder after the quiz; (j)~example photo gallery; (k--m)~the three quiz pairs, each answered
yes or no.}
\label{fig:study_materials}
\end{figure*}

This section reproduces the study as participants saw it. \cref{fig:study_materials} shows
the pages of the video study (a--c) and of the photo pilot (d--m), and the consent forms are
given verbatim at the end of this section.

\noindent\textbf{Recruitment and eligibility.} Participants were recruited
through Prolific. Eligibility required being 18 or older, fluent in English,
located in the United States, and using a mobile device with a camera. The
study was reviewed by our Institutional Review Board and exempted from
further oversight. Participants consented before beginning and were free to
withdraw at any point.

\noindent\textbf{Task.} Each video participant was shown a demonstration video of
an author filming a box of text while walking backwards
(\cref{fig:study_materials}a), then asked to record six videos of different texts in the
same way: start close with the text clearly readable, start filming, and take 5--10 steps
backwards (\cref{fig:study_materials}b). Participants in the pilot instead took 25 pairs of
photographs of the same text, one close-up and one far away or at a side angle
(\cref{fig:study_materials}d). They were shown correct and incorrect example pairs
(\cref{fig:study_materials}e,f) and had to judge three pairs in a short quiz before
starting (\cref{fig:study_materials}k--m). Both tasks asked for text with ten or more words.

\noindent\textbf{Privacy instructions.} Participants were told not to film
anything with personal information, and were given examples to exclude:
mail, addresses, usernames and e-mails, faces, uniquely identifying names, prescription
bottles, and license plates. Photo participants confirmed this with a checkbox
(\cref{fig:study_materials}g). Submissions were screened against these instructions before
entering \dataset.

\noindent\textbf{Compensation and sample.} The median completion time was
12:05 minutes at an average rate of \$9.93 per hour. The final approved
sample was 244 participants, skewed towards middle-aged participants
(median 37.68 years, range 18--76, SD 10.54) and roughly balanced by gender
(111 men, 130 women, 2 other), with a range of ethnicities, education levels
and employment statuses. The study cost \$818.34 including pilots.

\noindent\textbf{Release.} \dataset is released under the gated-access terms
in \cref{sec:discussion}: institutional verification and a Data Use
Agreement prohibiting de-anonymization.

\noindent\textbf{Consent forms.} Reproduced verbatim; the section headings are shortened.

{\footnotesize
\noindent\emph{Video study.} \textsc{Description.} The purpose of this research is to train an
AI to read text. We would like you take videos of text. You will read instructions, take 6
short videos, and upload them. The study should take roughly 11 minutes to complete.
\textsc{Risks.} There is a potential risk of a confidentiality breach. To minimize your risk,
you should avoid taking videos of any personal information. Only take videos of things that
you would feel comfortable uploading onto the internet. This video data will be made publicly
available as a dataset on the Web. We may also use this data for future research projects, and
the results will be presented in a published research paper and presentation. The videos you
upload will be processed by OpenAI via their API, but will not be stored or used by OpenAI.
\textsc{Benefits.} There are no direct benefits to you. \textsc{Compensation.} You will
receive \$2.00 for participating in this study. The study should take approximately 11
minutes to complete (\$10.91/hr). We reserve the right to refuse compensation if you did not
upload any videos. \textsc{Confidentiality.} We will take all necessary measures to respect
your privacy and maintain confidentiality while collecting, analyzing, and presenting data.
Any identifiable information in videos will be removed before being made publicly available.
\textsc{Questions.} If you have questions about the research, please contact the principal
investigator. Your participation is completely voluntary and can be withdrawn at any time.

\noindent\emph{Photo pilot.} \textsc{Description.} The purpose of this research is to create
an AI model that is capable of reading text that is very small, blurry, or angled. We would
like you take photos of text. This research will include participants who are
English-speaking, U.S.-based, and 18+ years old. You will read instructions, finish a short
quiz, and be asked to take and upload 25 pairs of photos. The study should take approximately
25 minutes to complete. \textsc{Risks.} There is a potential risk of a confidentiality breach.
We will be publicizing this data along with our research paper and AI model. To minimize your
risk, you should avoid taking photos of any personal information (e.g., names, addresses,
usernames, faces, mail, etc.). Any personal information we detect will be deleted. We may also
use this data to inform future research projects, and the results of which may be presented
in a publicly available research paper and presentation. The photos you upload will be
processed by OpenAI via their API, but will not be stored or used by OpenAI.
\textsc{Benefits.} There are no direct benefits to you. \textsc{Compensation.} You will
receive \$5 for participating in this study. The study should take approximately 25 minutes
to complete. We reserve the right to refuse compensation if we find that you did not follow
the instructions and took incorrect photo pairs or photos of personal information.
\textsc{Confidentiality.} We will take all necessary measures to respect your privacy and
maintain confidentiality while collecting, analyzing, and presenting data. Any identifiable
information be deleted. Photos will be aggregated and anonymized prior to publishing.
\textsc{Questions.} If you have questions about the research, please contact the principal
investigator. Your participation is completely voluntary and can be withdrawn at any time.
\par}

\section{Training Data Composition}
\label{sec:appendix_training_data}

\cref{tab:dataset_breakdown} lists the training and evaluation data. \name is
trained on \dataset video frames only: 6,734 frames from 1,133 videos at the six
sampled positions, labeled by the automatic pipeline, each with its own
\dataset instruction. Training shares no videos with the validation and test sets.
We tried two additions on the validation split and dropped both. The automatic \dataset
photo labels are truncated (43 words on average against 93 in the human-verified labels)
and taught the model to stop early, and the public OCR and QA sets gave no gain.

The recipe is supervised fine-tuning followed by GRPO with a BoW F1 reward, with every
image upscaled $\times$3 (bicubic) for training and inference. For SFT the 4B and 8B models are
fully fine-tuned (learning rates $10^{-6}$ language model, $10^{-5}$ projector, $10^{-6}$
vision encoder; 420 steps, effective batch 32, cosine schedule), and we keep the checkpoint
with the lowest validation loss (step 400). GRPO adds a fresh rank-64 LoRA, samples 8
completions per frame, and uses BoW F1 as the reward with a KL penalty of 0.02 against the
SFT model, for 300 steps.

\begin{table}[t]
\centering
\caption{Training and evaluation data. \name is fine-tuned on \dataset video frames only;
\dataset photos and the public sets were tried on validation and excluded
(\cref{sec:appendix_training_data}), so every other row is evaluation-only. \dataset test rows
are the held-out, human-verified split; a further 369 validation images (333 frames from 56 videos,
36 photos; 372 prompt rows) were used only to select the recipe and checkpoints.}
\label{tab:dataset_breakdown}
\footnotesize
\setlength{\tabcolsep}{3pt}
\begin{tabular}{l|l|l|c|c}
\toprule
\textbf{Category} & \textbf{Dataset} & \textbf{Primary Task} & \textbf{Train} & \textbf{Test} \\
\midrule
{Scene Text} & HierText \cite{long2022towards} & Hierarchical Text & -- & 1,000 \\
& Total-Text \cite{CK2019} & Curved Text & -- & 255 \\
& COCO-Text \cite{veit2016coco} & Scene Text Detection & -- & 500 \\
& ICDAR \cite{karatzas2015icdar} & In-the-wild Text & -- & 200 \\
& TextZoom \cite{wang2020scene} & Super-Resolution & -- & 67 \\
& Occluded \cite{weinman2024counting} & Occlusion Robustness & -- & 22 \\
\midrule
{Handwriting} & IAM-Line \cite{marti2002iam} & English Cursive & -- & 200 \\
& RIMES \cite{grosicki12008rimes} & French Handwriting & -- & 399 \\
& GNHK \cite{lee2021gnhk} & Handwritten Notes & -- & 65 \\
& Maptext \cite{li2024icdar} & Historical Maps & -- & 200 \\
\midrule
{\dataset} & Video \dataset & Noisy Text & 6,734 & 1,331 \\
& Photo \dataset & Noisy Text (Pilot) & -- & 241 \\
\midrule
\textbf{Total} & & & \textbf{6,734} & \textbf{4,480} \\
\bottomrule
\end{tabular}
\end{table}

\section{Ground-Truth Verification}
\label{sec:appendix_golden}

The test set is the human-verified golden set of \cref{sec:data}. Every unit has two
independent annotations: A1 from the automatic GPT-5 annotator and A2 from a human expert
annotator (\cref{tab:golden_composition}). A video unit is labeled once and applied to
its 5--6 sampled frames, so the 280 video units give 1,664 evaluation items; the 2 flagged
videos are excluded. Annotation took 12.1\,h for the videos and 25.8\,h for the photos.

\cref{tab:interannotator_agreement} reports the agreement between A1 and A2. On video the
automatic labels are close to the human ones (80.0\% exact agreement, mean WER 0.065), which
suggests the automatic video labels used for training are close to human quality. Photos agree
far less (43.7\%, mean WER 0.491). \cref{tab:completeness} shows that this disagreement is
almost entirely omission: no unit in either split was shortened by more than a factor of three
under A2, while 101 of the 277 photo units more than tripled in length. A1 under-transcribes
dense and non-Latin text rather than inventing text, which is why the automatic photo labels
were left out of training (\cref{sec:appendix_training_data}).

\begin{table}[t]
\centering
\caption{Composition of the human-verified BSTRB test set. A \textsc{res\_video}
unit is one video, annotated once for the whole video and applied to its 5--6 sampled
frames; a \textsc{res\_photo} unit is one image. Every unit carries two independent
annotations: A1 from the automatic GPT-5 annotator and A2 from the human expert
annotator, who worked from the frames without editing A1's output in place.}
\label{tab:golden_composition}
\begin{tabular}{l S[table-format=3.0] S[table-format=3.0]}
\toprule
& {\textsc{res\_video}} & {\textsc{res\_photo}} \\
\midrule
Units annotated            & 282 & 277 \\
\quad accepted unchanged   & 222 & 119 \\
\quad edited               & 58 & 158 \\
\quad flagged (excluded)   & 2 & 0 \\
\midrule
Units in benchmark         & 280 & 277 \\
Evaluation items           & 1664 & 277 \\
\midrule
Annotation time (h)        & 12.1 & 25.8 \\
Median time per unit (s)   & 23 & 17 \\
\bottomrule
\end{tabular}
\end{table}

\begin{table*}[t]
\centering
\caption{Inter-annotator agreement on BSTRB between A1, the automatic GPT-5
annotator, and A2, the human expert annotator. Agreement is computed after the
normalisation used for all BSTRB scoring. CER and WER are directional, taking A1 as
hypothesis and A2 as reference, and are clipped at $1.0$ so that units whose A1 label is
near-empty cannot dominate the mean; \emph{Agr.} is the symmetric normalised edit
similarity $1 - d(\mathrm{A1},\mathrm{A2})/\max(|\mathrm{A1}|,|\mathrm{A2}|)$,
which does not depend on which annotator is treated as the reference.}
\label{tab:interannotator_agreement}
\begin{tabular}{l S[table-format=3.1] S[table-format=1.3] S[table-format=1.3] S[table-format=1.3] S[table-format=1.3] S[table-format=1.3]}
\toprule
& {Exact agr.\ (\%)} & {CER mean} & {CER med.} & {WER mean} & {WER med.} & {Agr.\ mean} \\
\midrule
\textsc{res\_video} & 80.0 & 0.058 & 0.000 & 0.065 & 0.000 & 0.961 \\
\textsc{res\_photo} & 43.7 & 0.481 & 0.333 & 0.491 & 0.393 & 0.583 \\
\bottomrule
\end{tabular}
\end{table*}

\begin{table}[t]
\centering
\caption{Transcript length under A1 (automatic) against A2 (human). Disagreement is
near-exclusively \emph{additive}: no unit in either split was shortened by more than a
factor of three under A2, while 101 of 277 \textsc{res\_photo}
units more than tripled in length. A1 under-transcribed dense and non-Latin text rather
than hallucinating text that was not present, so the disagreement is dominated by
omission rather than by substitution.}
\label{tab:completeness}
\begin{tabular}{l S[table-format=3.0] S[table-format=3.0]}
\toprule
& {\textsc{res\_video}} & {\textsc{res\_photo}} \\
\midrule
Median A1 transcript (chars)   & 242 & 102 \\
Median A2 transcript (chars)   & 246 & 342 \\
\midrule
Units with A1 $<$30 chars                & 13 & 73 \\
Units where A2 $>3\times$ A1             & 1 & 101 \\
Units where A2 $<\frac{1}{3}$ A1     & 0 & 0 \\
\bottomrule
\end{tabular}
\end{table}

\section{\dataset Degradation Characteristics}
\label{sec:appendix_dataset}

\begin{figure*}[tp]
\centering
\begin{subfigure}[b]{0.66\textwidth}
  \includegraphics[width=\textwidth]{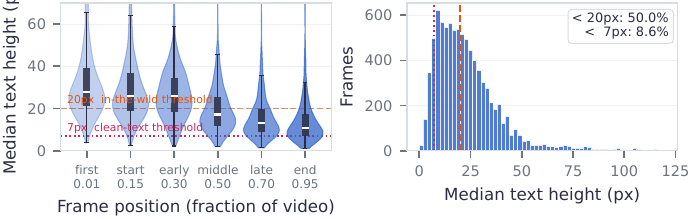}
  \caption{Median text height by frame position, and pooled.}
  \label{fig:text_height_dist}
\end{subfigure}

\vspace{4pt}
\begin{subfigure}[b]{0.66\textwidth}
  \includegraphics[width=\textwidth]{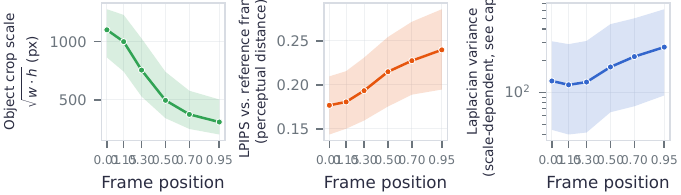}
  \caption{Object crop scale, LPIPS to the reference frame, and Laplacian
  variance across the walk-back.}
  \label{fig:degradation_curves}
\end{subfigure}

\vspace{4pt}
\begin{subfigure}[b]{0.36\textwidth}
  \includegraphics[width=\textwidth]{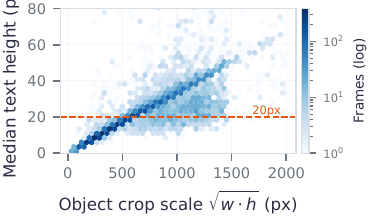}
  \caption{Joint coverage of object scale and text height.}
  \label{fig:scale_vs_height}
\end{subfigure}
\caption{\dataset degradation characteristics over 8,889 per-frame records.
Laplacian variance rises only because it is scale-dependent.}
\label{fig:appendix_dataset}
\end{figure*}

\cref{fig:appendix_dataset} characterizes the degradation in \dataset over the 8,889
per-frame records with complete degradation metadata, after removing the 13 units flagged
in our PII audit.

\noindent\textbf{Text height spans both thresholds.}
Median text height falls from 28.0px in the reference frame to 10.9px in the
final frame (\cref{fig:text_height_dist}). Pooled over all frames, 50.0\% are below the
20px in-the-wild threshold and 8.6\% below 7px.

\noindent\textbf{Degradation is compound, not just downscaling.}
Object crop scale falls from 1,100px to 311px while LPIPS distance to the
reference frame rises from 0.177 to 0.240 (\cref{fig:degradation_curves}). Laplacian
variance \emph{rises} over the same sweep, but because it is scale-dependent, this does not
mean the later frames are sharper.

\noindent\textbf{Scale and text height are jointly covered.}
Prior datasets provide either a scale sweep or fine-grained spatial
annotation. The joint coverage in \cref{fig:scale_vs_height} is what allows
performance to be plotted as a continuous function of text pixel size in
\cref{fig:bstrb}.

\noindent\textbf{Text heights on late frames are derived, not measured.}
Per \cref{sec:data}, word boxes are detected by OCR on the sharp reference
frame and scaled by the SAM3 object bounds as the object recedes, so heights
on frames 3--6 are propagated rather than re-detected. They are consistent
with \cref{eq:height} (log-space $r=0.75$), but have not been checked against
heights detected on the later frames themselves.

\noindent\textbf{No leakage.} The 1,133 training videos, the 56 validation videos and the
224 test videos share no \texttt{base\_id}. Evaluation uses the human-verified test split
(1,331 video frames and 241 photos).

\begin{figure*}[tp]
\centering
\includegraphics[width=\textwidth]{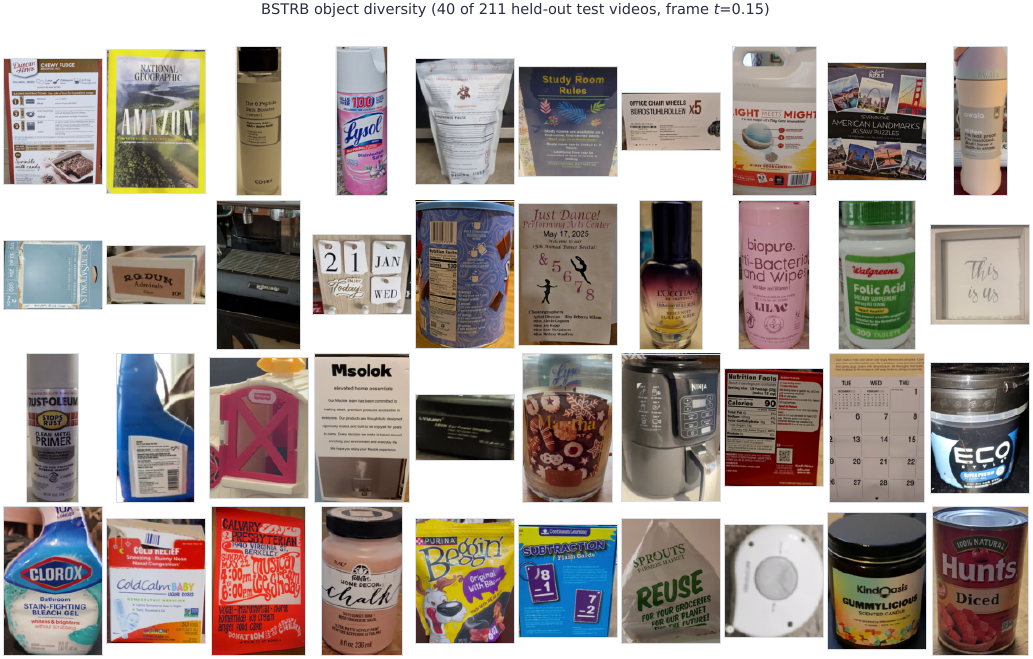}
\caption{40 of the 211 held-out test clips at $t{=}0.15$, one
frame per object. Note the variance in text density and style.}
\label{fig:dataset_diversity}
\end{figure*}

\section{Evaluation Protocol}
\label{sec:appendix_protocol}

\noindent\textbf{Prompts.} One prompt per sample, drawn at random from a
per-dataset pool (34 for \dataset-video, 24 for RIMES, 11 for COCO-Text).
Pools contain both transcription and captioning instructions; we score the
transcription register unless stated otherwise.

\noindent\textbf{System prompt.} For the transcription track and \dataset, every model
receives the same transcription-only system prompt (output the text and nothing else; output
nothing if no text is legible), followed by the item's own prompt and the image. The OOD and
Snellen evaluations use each item's prompt only.

\noindent\textbf{Decoding.} Local models (\name and the Qwen3-VL bases) decode greedily with
at most 4,096 new tokens. \name's input is upscaled $\times$3 (bicubic) before the processor's
1{,}003{,}520-pixel cap; the bases are run without upscaling. API models use temperature 0
where the endpoint accepts it and the provider default otherwise (reasoning effort at the
provider default), with at most 16,000 output tokens: 32,768 for the reasoning models (GPT-5,
GPT-5.6, GPT-6, o3, o4-mini), 65,536 for Kimi K2.6 and 8,192 for Llama~4. Outputs that reach the
cap are scored as produced.

\noindent\textbf{Models.} 20 models are served through a single LiteLLM proxy (Azure and AWS
Bedrock deployments): GPT-4o, GPT-4o mini, GPT-4.1, GPT-5, GPT-5.4, GPT-5.4 mini, GPT-5.4 nano,
GPT-5.6 Sol, Terra and Luna, GPT-6 Astra, o3, o4-mini, Kimi K2.5 and K2.6, Qwen3-VL-235B-A22B,
Llama~4 Maverick and Scout, Gemma~3 4B and 12B. Each result records the deployment's
\texttt{model\_id}. Anthropic models were excluded because they decline to reproduce the long
copyrighted book text in the test set.

\noindent\textbf{Scoring.} Lowercase, punctuation-stripped, whitespace-split, no stemming, on
the raw outputs. Stripping markdown, preambles and bullets first (\cref{fig:heatmap_clean})
changes the score of 3--5\% of each model's outputs and moves no model's mean by more than
0.07\,pp.

\section{Additional Benchmark Results}
\label{sec:appendix_benchmarks}

\begin{figure*}[tp]
\centering
\includegraphics[width=\textwidth]{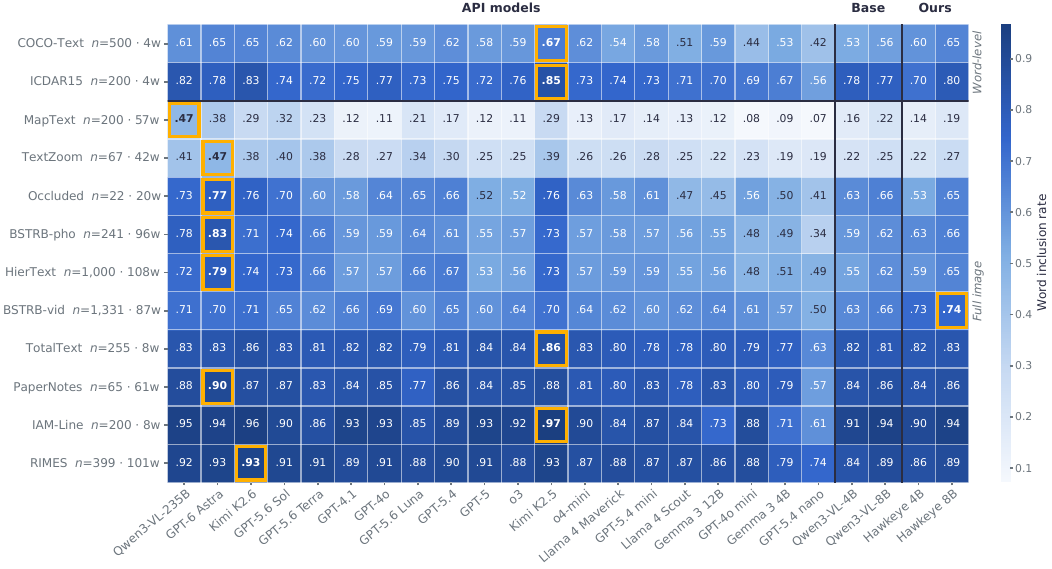}
\caption{Transcription track, word inclusion rate; layout as in \cref{fig:heatmap_f1}.
GPT-6 Astra is best on 5 of 12 columns; Hawkeye (Qwen3-VL-8B) is best on \dataset-vid. The best model per column is the same as under BoW F1 on
8 of 12 columns.}
\label{fig:heatmap_wir}
\end{figure*}

\cref{fig:heatmap_wir} repeats the transcription track under word inclusion rate. The best
model per dataset differs from BoW F1 on 4 of the 12 datasets: GPT-6 Astra leads 5 datasets,
Kimi K2.5 4, and Kimi K2.6, Qwen3-VL-235B and \name 8B one each. \name 8B leads on \dataset
video, as it does under BoW F1. WIR is a pure recall term and so favors models that emit more
text, whereas BoW F1 penalizes the words they add.

\begin{figure*}[tp]
\centering
\includegraphics[width=\textwidth]{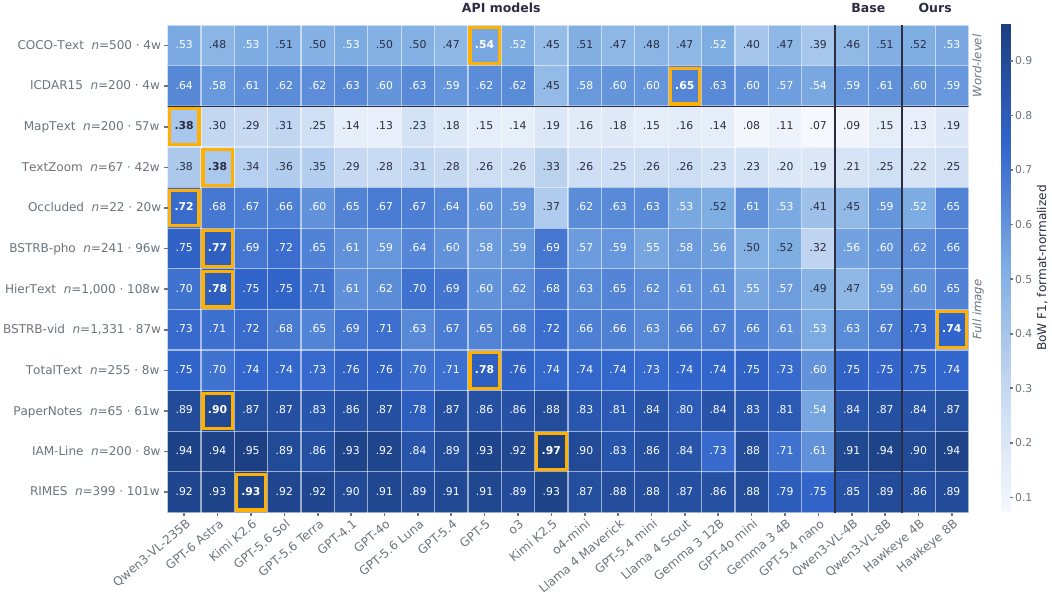}
\caption{\cref{fig:heatmap_f1} recomputed with output formatting normalized (markdown,
preamble and bullets stripped identically for every model). No API request failed, so no
sample is dropped. Normalization changes the score of 3--5\%
of each model's outputs and moves no model's mean by more than 0.07pp; the best
model per column is unchanged on 12 of 12 columns.}
\label{fig:heatmap_clean}
\end{figure*}

\cref{fig:heatmap_clean} applies a formatting check to \cref{fig:heatmap_f1}: markdown,
preambles and bullets are stripped identically for every model before scoring. No API request
failed, so no sample is dropped. With the shared transcription-only system prompt the models
rarely format their output, and normalization leaves the best model unchanged on all 12
datasets.

\begin{figure}[t]
\centering
\includegraphics[width=0.92\columnwidth]{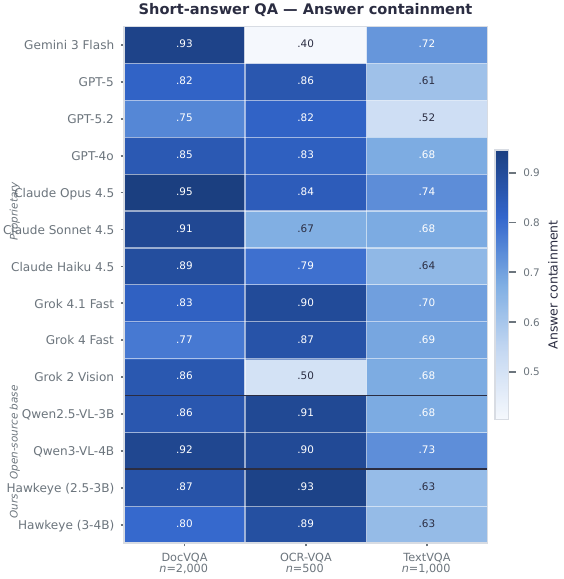}
\caption{Short-answer QA track scored with answer containment, for the preliminary 4B model
(\cref{sec:architecture}) and API models; it was not run for the final models.}
\label{fig:qa_heatmap}
\end{figure}

\cref{fig:qa_heatmap} gives a short-answer QA track, scored with answer containment, for the
preliminary 4B model of \cref{sec:architecture} and the API models; we did not run it for the
final models.

\begin{figure*}[tp]
\centering
\begin{subfigure}[b]{0.52\textwidth}
  \includegraphics[width=\textwidth]{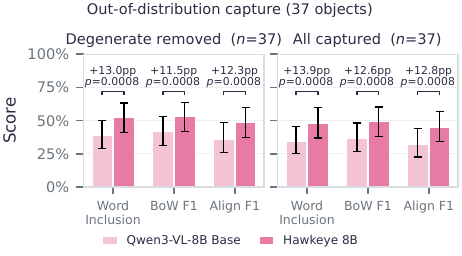}
  \caption{Aggregate, filtered and unfiltered}
  \label{fig:ood_metrics}
\end{subfigure}\hfill
\begin{subfigure}[b]{0.45\textwidth}
  \includegraphics[width=\textwidth]{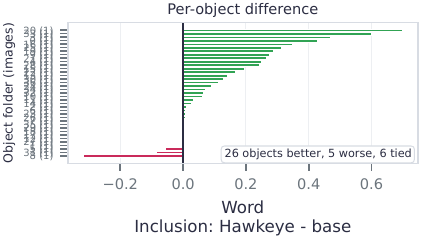}
  \caption{Per-object differences}
  \label{fig:ood_by_folder}
\end{subfigure}
\caption{37 objects captured with smart glasses outside the \dataset
distribution (screens, signage, documents, outdoors, lab, Chinese
characters): Hawkeye 8B against its base Qwen3-VL-8B.}
\label{fig:ood}
\end{figure*}

\noindent\textbf{Out of distribution.} \cref{fig:ood} reports the smart-glasses captures of
\cref{sec:ood} (91 images of 37 objects) for \name 8B against its base Qwen3-VL-8B, paired over
images and Holm-corrected. \name 8B gains on every metric, with and without the
degenerate-output filter (\cref{fig:ood_metrics}): word inclusion $+13.0$\,pp
($[+6.9, +19.7]$), BoW F1 $+11.5$\,pp and aligned F1 $+12.3$\,pp (all
$p_{\mathrm{Holm}} < 0.001$). Per object, 26 improve and 5 regress (\cref{fig:ood_by_folder}).
\name 4B also gains in BoW F1 ($+14.7$\,pp) and aligned F1 ($+16.6$\,pp), but not
significantly in word inclusion ($+6.4$\,pp).

\noindent\textbf{Why we cluster.} \dataset frames are grouped by source video (each video
contributes up to six frames that share one transcript), so the effective sample size is the
number of videos, not the number of frames. Treating frames as independent understates the
standard error of mean \dataset-video BoW F1 by $1.56$--$1.73\times$ across all 24 models, enough
to turn a null result into a significant one. For Snellen, most variance is between optotype
sizes within a chart, so chart-level clustering does not inflate the error in the same way.

\section{Error Composition}
\label{sec:appendix_errors}

\begin{figure*}[tp]
\centering
\begin{subfigure}[b]{0.6\textwidth}
  \includegraphics[width=\textwidth]{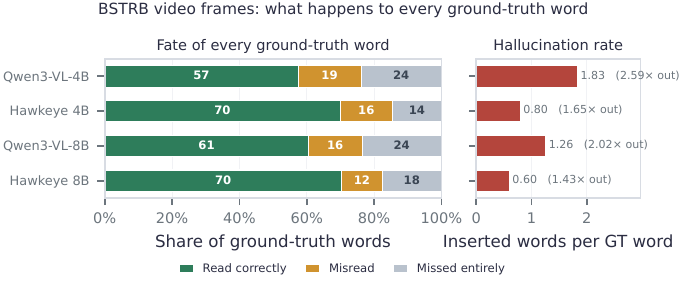}
  \caption{Every ground-truth word on held-out \dataset video frames assigned to one of
  correct, misread or missed (left), and inserted words per ground-truth word (right).}
  \label{fig:error_taxonomy}
\end{subfigure}\hfill
\begin{subfigure}[b]{0.37\textwidth}
  \includegraphics[width=\textwidth]{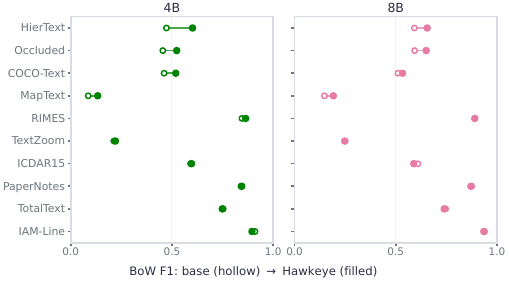}
  \caption{The ten general OCR sets before and after fine-tuning.}
  \label{fig:forgetting}
\end{subfigure}
\caption{Error composition and what fine-tuning costs, for Hawkeye at each size and its
Qwen3-VL base.}
\label{fig:appendix_errors}
\end{figure*}

\cref{fig:appendix_errors} breaks down the errors on the \dataset test video frames.
\cref{fig:error_taxonomy} assigns every ground-truth word to one of four outcomes (correct,
misread, missed, or inserted), which separates reading more from guessing more. At both sizes
fine-tuning converts misses into correct reads while emitting less text: correct words rise
from 57\% to 70\% (4B) and from 61\% to 70\% (8B), missed words fall from 24\% to 15\% and from
24\% to 18\%, and inserted words per ground-truth word fall from 1.83 to 0.80 and from 1.26 to
0.60. Output length drops from $2.6\times$ to $1.7\times$ and from $2.0\times$ to $1.4\times$
the reference.

\cref{fig:forgetting} shows the ten general OCR sets before and after fine-tuning (paired tests
in \cref{fig:forest_base}). No set regresses significantly at either size. The gains are on
HierText, MapText and, for 4B, COCO-Text and RIMES; the other sets are within a few points of
the base.

\begin{figure*}[tp]
\centering
\includegraphics[width=\textwidth]{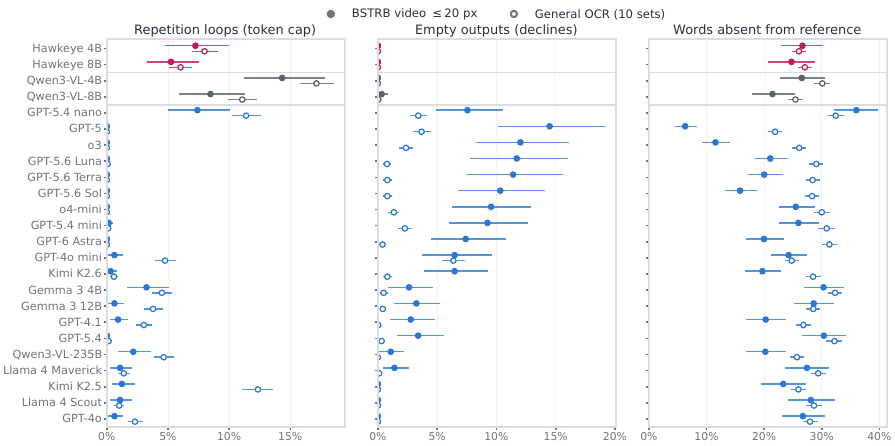}
\caption{How each model fails, on held-out \dataset video frames at $\leq$20\,px (filled) and
the 10 general OCR sets (open), 95\% CIs clustered by video. Left: outputs that ran to the token
cap (repetition loops). Middle: empty outputs. Right: over the remaining outputs, the share of
words absent from the reference, an upper bound on hallucination. At $\leq$20\,px, Hawkeye 4B
loops on 7.2\% of frames against 14.3\%
for its base, and Hawkeye 8B on 5.2\% against
8.5\%. Once loops are removed, Hawkeye's unsupported-word share differs from its base's by 4B +0.1, 8B +3.3\,pp (intervals overlap at every size). API models rarely loop but return nothing instead, up to
14.5\% of frames (GPT-5).}
\label{fig:hallucination}
\end{figure*}

\noindent\textbf{Hallucination and failure modes.} The inserted-word rate above pools every
output, so a few runaway repetition loops dominate it. \cref{fig:hallucination} and
\cref{tab:hallucination} separate three ways a model can fail: looping to the token cap,
returning nothing, and emitting words that are not in the reference. Most of fine-tuning's
effect is on the loops: on \dataset video at $\leq$20\,px, \name 4B loops on 7.2\% of frames
against 14.3\% for its base, and \name 8B on 5.2\% against 8.5\%. Once loops are removed, the
share of output words absent from the reference is close to the base model's at both sizes
($+0.1$ and $+3.3$\,pp; intervals overlap), so fine-tuning does not make individual outputs
hallucinate less; it mostly stops them looping. The API models rarely loop, but they decline
on hard frames, returning nothing on up to 14.5\% of $\leq$20\,px frames (GPT-5; GPT-6 Astra
7.4\%, Kimi K2.6 6.5\%).

\begin{table*}[t]
\centering
\caption{Hallucination and output failures. Loop: outputs that ran to the token cap. Over the
remaining non-empty outputs, Uns.: mean share of output words absent from the reference (an upper
bound on hallucination), and Ins.: mean Levenshtein insertions per reference word. Hawkeye and its base at each size, and the three API models with the
fewest unsupported words on \dataset video $\leq$20\,px. All models: \cref{fig:hallucination}.}
\label{tab:hallucination}
\scriptsize
\setlength{\tabcolsep}{3pt}
\begin{tabular}{lrrrrrrrrrrrr}
\toprule
 & \multicolumn{3}{c}{BSTRB video $\leq$20\,px} & \multicolumn{3}{c}{BSTRB video $>$20\,px} & \multicolumn{3}{c}{BSTRB photo} & \multicolumn{3}{c}{General OCR (10 sets)} \\
 & Uns.\,\% & Ins. & Loop\,\% & Uns.\,\% & Ins. & Loop\,\% & Uns.\,\% & Ins. & Loop\,\% & Uns.\,\% & Ins. & Loop\,\% \\
\midrule
Qwen3-VL-4B & 26.5 & 0.10 & 14.3 & 13.9 & 0.13 & 5.9 & 29.9 & 0.15 & 11.6 & 30.1 & 0.37 & 17.1 \\
Hawkeye 4B & 26.6 & 0.10 & 7.2 & 12.6 & 0.11 & 2.2 & 26.2 & 0.18 & 9.1 & 26.0 & 0.43 & 8.0 \\
\midrule
Qwen3-VL-8B & 21.4 & 0.07 & 8.5 & 10.7 & 0.11 & 4.1 & 20.1 & 0.12 & 9.5 & 25.4 & 0.39 & 11.1 \\
Hawkeye 8B & 24.7 & 0.10 & 5.2 & 10.8 & 0.11 & 1.9 & 24.3 & 0.23 & 5.0 & 27.0 & 0.53 & 6.0 \\
\midrule
GPT-5 & 6.3 & 0.04 & 0.0 & 5.9 & 0.04 & 0.0 & 17.7 & 0.04 & 0.0 & 21.9 & 0.16 & 0.0 \\
o3 & 11.5 & 0.06 & 0.0 & 8.1 & 0.06 & 0.0 & 20.2 & 0.10 & 0.0 & 26.1 & 0.28 & 0.0 \\
GPT-5.6 Sol & 15.8 & 0.06 & 0.0 & 9.9 & 0.09 & 0.0 & 21.0 & 0.27 & 0.0 & 28.3 & 0.38 & 0.0 \\
\bottomrule
\end{tabular}
\end{table*}

\section{Attention and Probing}
\label{sec:appendix_attn}

We recorded each generated token's attention to the image patches (last four layers, averaged
over heads) for \name 4B and 8B and their Qwen3-VL bases on 160 \dataset test video frames
from 125 videos, each with its own prompt and up to 256 generated tokens. Within each pair both
arms see the same input, the frame upscaled $\times$3 and capped at $10^6$ pixels (a median of 944
visual tokens), so neither has a resolution advantage. The unit of analysis is the image:
per-token values are averaged per image, base and \name are paired over images, and we report a
bootstrap 95\% CI of the paired difference and the standardized effect $d_z$.

\noindent\textbf{Attention sinks.} About two to three content-free patches per image (mean
attention over 50 times the median patch) hold 41\% (4B) and 32\% (8B) of the base models'
attention and, in the median image, are the argmax for 98--100\% of tokens. A peak or top-$k$
measure computed on raw maps would therefore describe the sink, not reading, so we report the
mass on these patches separately and exclude them from every other measure.

\begin{figure*}[tp]
\centering
\includegraphics[width=0.9\textwidth]{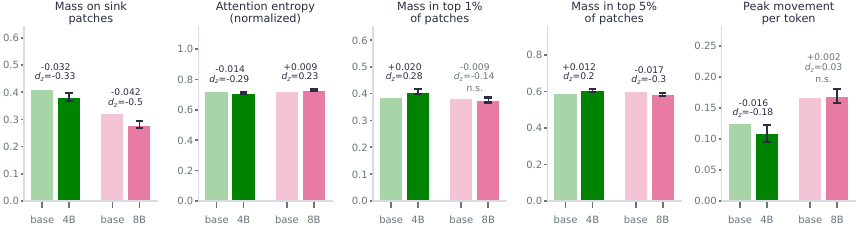}
\caption{Attention of Hawkeye and its Qwen3-VL base at 4B (160 frames) and 8B (159 frames) on held-out
\dataset video frames, same input for both arms (bicubic $\times$3, 1\,MP cap), last four layers'
attention to image patches averaged over generated tokens; bars are image means (light: base,
dark: Hawkeye), whiskers the paired 95\% bootstrap CI of the difference. Content-free ``sink''
patches (mean attention $>$50$\times$ the median patch) are reported separately and excluded from
the other measures. Hawkeye $-$ base, 4B / 8B: sink mass -0.032 ($d_z$=-0.33) / -0.042 ($d_z$=-0.5);
entropy -0.014 ($d_z$=-0.29) / +0.009 ($d_z$=0.23); top-1\% mass +0.020 ($d_z$=0.28) /
-0.009 ($d_z$=-0.14, n.s.).}
\label{fig:attention_stats}
\end{figure*}

\begin{figure*}[tp]
\centering
\includegraphics[width=0.82\textwidth]{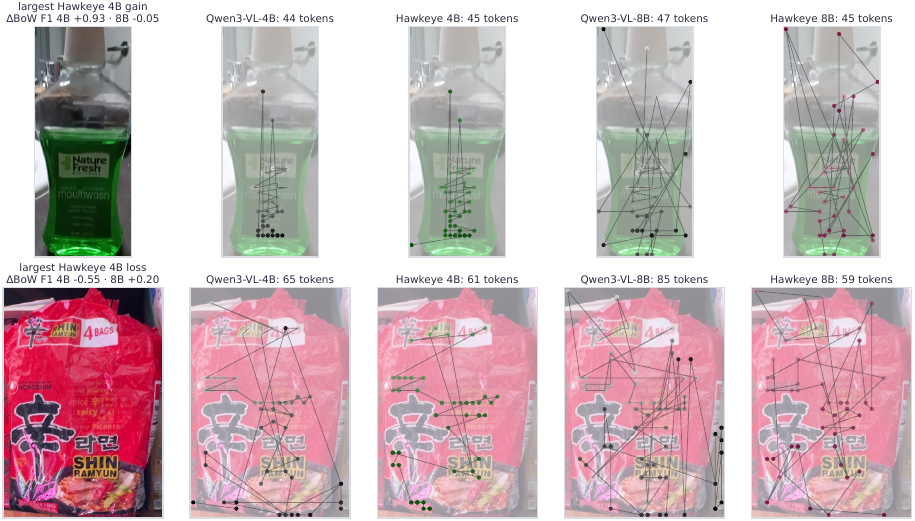}
\caption{Reading traces: the attention peak of each generated token (sink patches excluded),
darker for later tokens, for base and Hawkeye at 4B and 8B on the two analysed frames with the
largest Hawkeye 4B gain and loss in BoW F1 (selection rule fixed in advance; the same frames are
shown for 8B).}
\label{fig:attention_trace}
\end{figure*}

\noindent\textbf{Fine-tuning moves attention off the sinks} (\cref{fig:attention_stats}). At
both sizes \name puts less attention on the sink patches: $0.413 \rightarrow 0.381$ at 4B
($d_z{=}-0.33$) and $0.323 \rightarrow 0.281$ at 8B ($d_z{=}-0.50$). How the remaining attention
is distributed changes little and not in a consistent direction: at 4B it becomes slightly more
concentrated (entropy $-0.014$, $d_z{=}-0.29$; top-1\% mass $+0.020$, $d_z{=}0.28$), at 8B
slightly less (entropy $+0.009$, $d_z{=}0.23$; top-5\% mass $-0.017$, $d_z{=}-0.30$). We
therefore do not claim that fine-tuning concentrates or disperses attention: the preliminary
model appeared to disperse attention, but only when the sinks were not separated. \cref{fig:attention_trace} shows the per-token peaks on the two analyzed frames with
the largest \name 4B gain and loss in BoW F1, for both sizes: all four models scan the label
text, and the 8B traces wander more across the frame than the 4B ones.

\begin{figure}[!htb]
\centering
\includegraphics[width=\columnwidth]{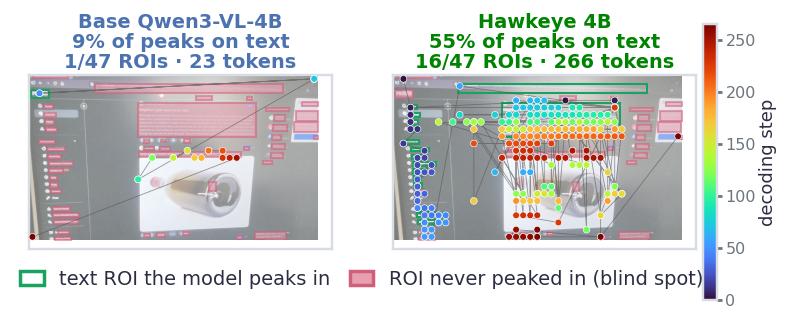}
\caption{Reading trace on a web page photographed with smart glasses: peak attention patch per
generated token, in decoding order (sinks excluded); green regions hold a peak, pink regions
none. Base Qwen3-VL-4B transcribes one post title; Hawkeye 4B reads the sidebar and the post
(BoW F1 .12 vs .84).}
\label{fig:blind_spots_manual}
\vspace{-4pt}
\end{figure}

\begin{figure*}[tp]
\centering
\begin{subfigure}[b]{0.495\textwidth}
  \includegraphics[width=\textwidth]{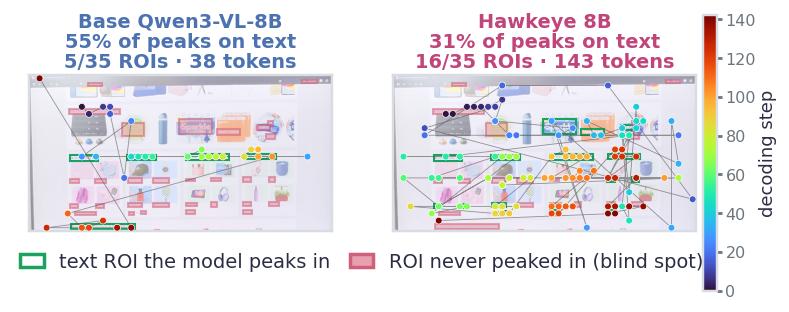}
  \caption{Amazon page: BoW F1 .44 (base) vs .82 (Hawkeye).}
\end{subfigure}\hfill
\begin{subfigure}[b]{0.495\textwidth}
  \includegraphics[width=\textwidth]{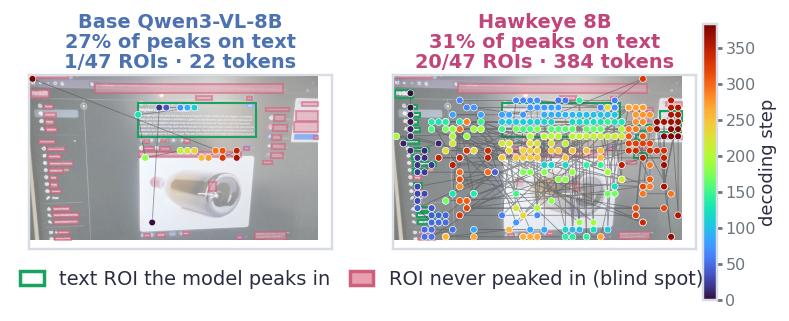}
  \caption{Reddit page: BoW F1 .13 (base) vs .93 (Hawkeye).}
\end{subfigure}
\caption{Reading traces of base Qwen3-VL-8B and Hawkeye 8B on two smart-glasses captures, same
conventions as \cref{fig:blind_spots_manual} (4B on the Reddit page). The base model lists the
section headings of the Amazon page (5 of 35 text regions) and one post title of the Reddit page
(1 of 47); Hawkeye reads the tile labels (16 regions) and the sidebar and post (20 regions).
Hand-picked among captures where Hawkeye's transcription beats its base's at both sizes;
over all captures the share of text regions visited does not differ (\cref{sec:appendix_attn}).
Regions: detector and OCR boxes merged into lines, plus tile labels added by hand.}
\label{fig:blind_spots_ebay}
\end{figure*}

\begin{figure*}[tp]
\centering
\includegraphics[width=\textwidth]{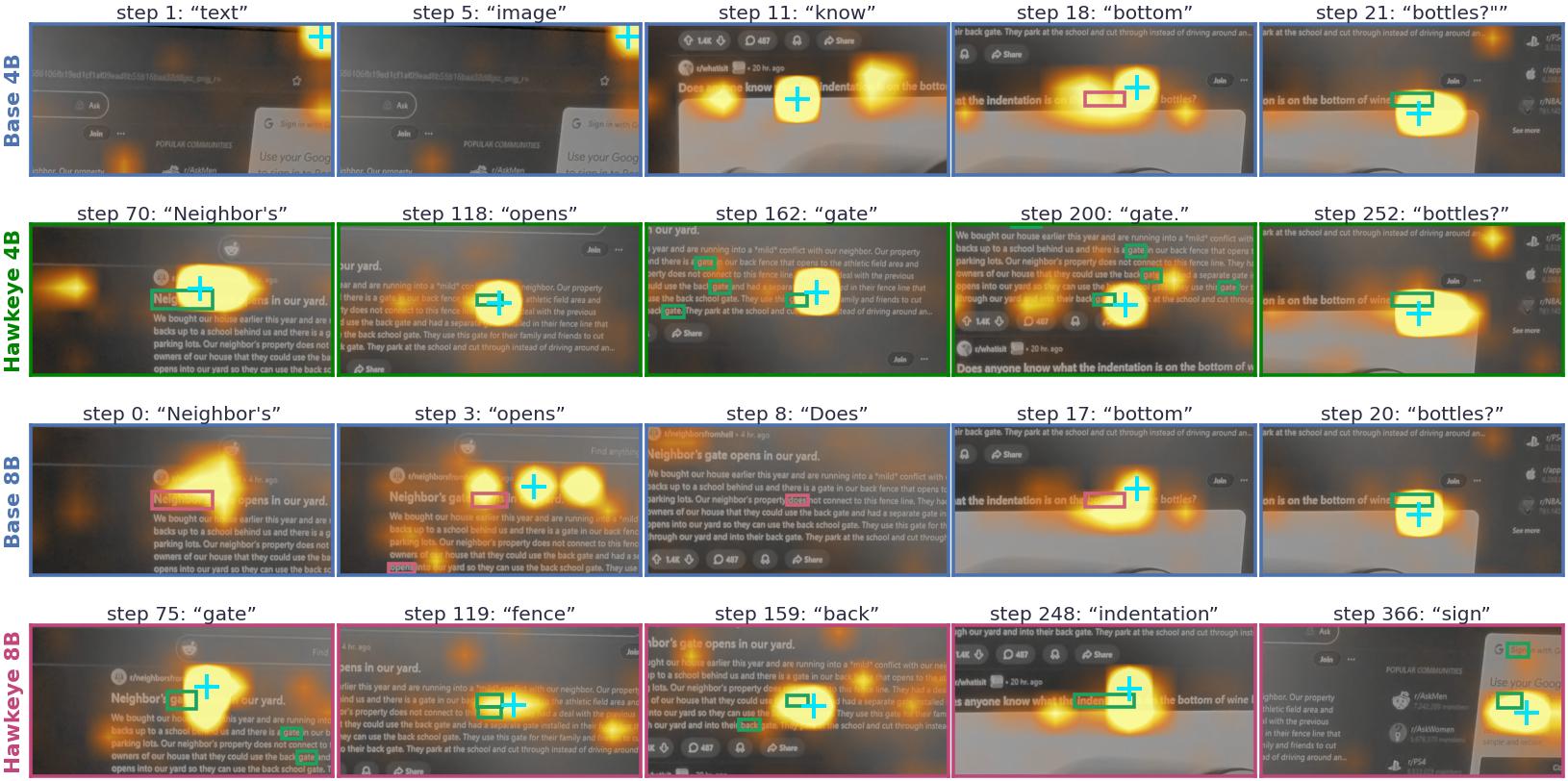}
\caption{Attention at individual decoding steps on the Reddit capture of
\cref{fig:blind_spots_manual,fig:blind_spots_ebay}. Each panel is the attention map (last four
layers, mean over heads, sinks excluded; brighter = more) at the step that emits the word in its
title, zoomed to the word; the cross marks the peak and the box the word as located by OCR
(green: peak on the word, pink: not). Words are content words in the model's own output, spread
over decoding and preferring those located by OCR. The bases spend their few steps on a
preamble and one title, with the peak often beside the word; Hawkeye's peak tracks the word it is
emitting through the post body.}
\label{fig:attention_filmstrip}
\end{figure*}

\noindent\textbf{Reading traces and word-level grounding.}
\cref{fig:blind_spots_manual,fig:blind_spots_ebay,fig:attention_filmstrip} are illustrations,
picked among the 91 smart-glasses captures of \cref{sec:ood} (up to 512 tokens) where \name's
transcription beats its base's by more than 0.1 BoW F1 at both sizes. On a Reddit page
photographed off a monitor (\cref{fig:blind_spots_manual}), both base models transcribe a single
post title (22--23 tokens, BoW F1 $0.12$ and $0.13$) and their peaks touch 1 of 47 text regions;
\name reads the sidebar, the post and the next title verbatim (BoW F1 $0.84$ at 4B, $0.93$ at
8B) and its trace covers 16 and 20 regions. On an Amazon page (\cref{fig:blind_spots_ebay}),
base Qwen3-VL-8B lists the section headings (38 tokens, BoW F1 $0.44$, 5 of 35 regions) while
\name 8B also reads the tile labels ($0.82$, 16 regions). Text regions are the detector and OCR
boxes merged into lines, plus tile labels the detector missed, added by hand. To check that a
trace reflects reading, \cref{fig:attention_filmstrip} shows the attention at the step that
emits a word: on the post body \name's peak sits on the word it is writing. Over all captures,
word-level grounding is the same for both models: for words that OCR also finds in the image,
the peak at the step emitting the word falls on that word $41\%$ vs $37\%$ of the time at 4B
(base vs \name) and $34\%$ vs $33\%$ at 8B, against about $2\%$ by chance (not significant).
Both models ground what they emit, but \name emits more of what is there (BoW F1 of these
outputs $+7.6$\,pp at 8B, $p_{\mathrm{Holm}} = 0.001$; $+5.1$\,pp at 4B, n.s.).

\noindent\textbf{Attention on detected text, at scale.} To go beyond single captures we detect
text regions automatically (the CRAFT detector of EasyOCR, without recognition, on the same
$\times$3 input; on two hand-annotated captures, 99\% and 90\% of its boxes fall inside the
annotated regions). On the 141 analyzed frames with at least one detected box (113 videos), we
measure the share of non-sink attention on text patches, the share of token peaks on text, and
the share of text boxes visited within a matched token budget. \name visits as many text boxes
as its base (base $\rightarrow$ \name: 4B $0.85 \rightarrow 0.84$, 8B $0.80 \rightarrow 0.80$,
n.s.), and the change in boxes visited does not predict the change in BoW F1 (Spearman
$|\rho| \le 0.09$). Its attention on detected text is lower ($-3.5$\,pp at 4B, $-3.4$\,pp at 8B,
$p_{\mathrm{Holm}} < 0.01$), but only on text at or below 20\,px ($-6.5$ and $-5.7$\,pp; above
20\,px $-1.0$ and $-1.5$\,pp). On small text the detector boxes the large print and misses the
fine print, and \name attends to that fine print and transcribes it (\cref{fig:showcase}).
Frames where \name's attention moves further off the detected text are also frames where it
outputs more words than its base (Spearman $\rho = -0.23$ at 4B, $p = 0.005$; $-0.16$ at 8B,
$p = 0.07$). On the 87 smart-glasses captures of \cref{sec:ood} with detected text (512 tokens),
coverage is again unchanged (boxes visited, base $\rightarrow$ \name: 4B
$0.63 \rightarrow 0.60$, 8B $0.62 \rightarrow 0.55$, n.s. after Holm), and \name again puts less
of its attention on the detected text (8B $-2.8$\,pp, $p_{\mathrm{Holm}} = 0.002$; 4B
$-1.3$\,pp, n.s.) and markedly less on the sinks ($-4.3$\,pp at 4B, $-6.4$\,pp at 8B); there
the shift is not tied to output length ($|\rho| \le 0.10$).

\section{Dataset and Output Examples}
\label{sec:appendix_examples}

\begin{figure*}[p]
\centering
\includegraphics[height=0.95\textheight]{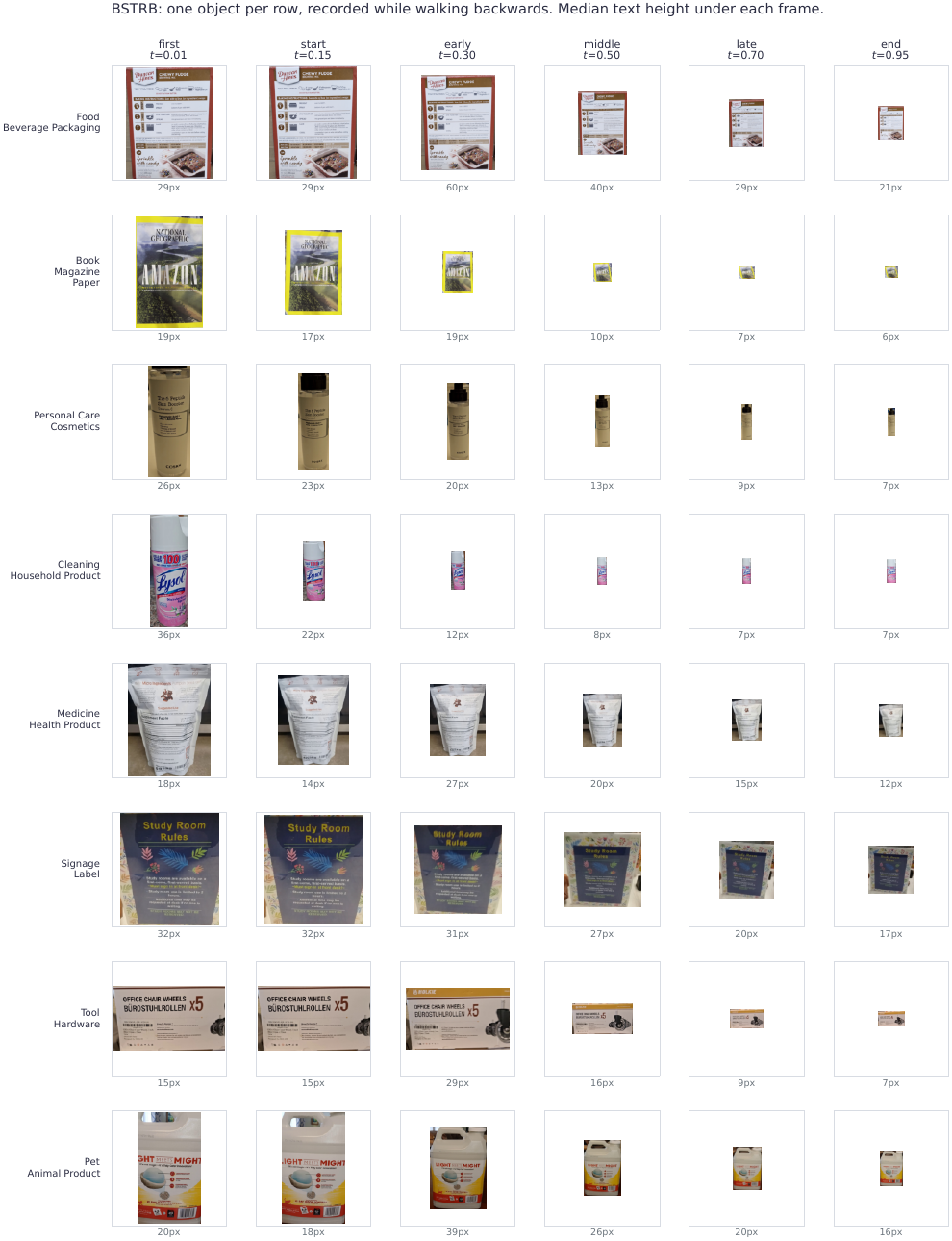}
\caption{\dataset examples, with median text height. Note, these examples in the paper are downsampled and not full-resolution.}
\label{fig:dataset_grid}
\end{figure*}

\cref{fig:dataset_grid} shows complete test clips: one object per row sampled at all
\nIntervals{} extraction intervals, annotated with the median text height of each frame. The
degradation is not a uniform blur. The object shrinks in the frame, so glyph evidence is lost
to scale rather than to defocus, and most rows are at or below the 20px threshold by
$t{=}0.70$. The fall-off is uneven: a book drops from 19px to 6px and a cleaning product from
36px to 7px across a single clip, while a sign stays at 17px or more. The first frame is not
always the largest either: a medicine box rises from 18px to 27px at $t{=}0.30$ as the
participant re-frames it before walking back.

\cref{fig:dataset_diversity} samples 40 of the 211 test clips at a single interval to show the
object distribution behind the category breakdown of \cref{fig:dataset_composition}: packaged
food, household chemicals, cosmetics, books, screens, and printed signage, photographed in
kitchens and living spaces under whatever lighting the participant had.

\begin{figure*}[p]
\centering
\includegraphics[height=0.95\textheight]{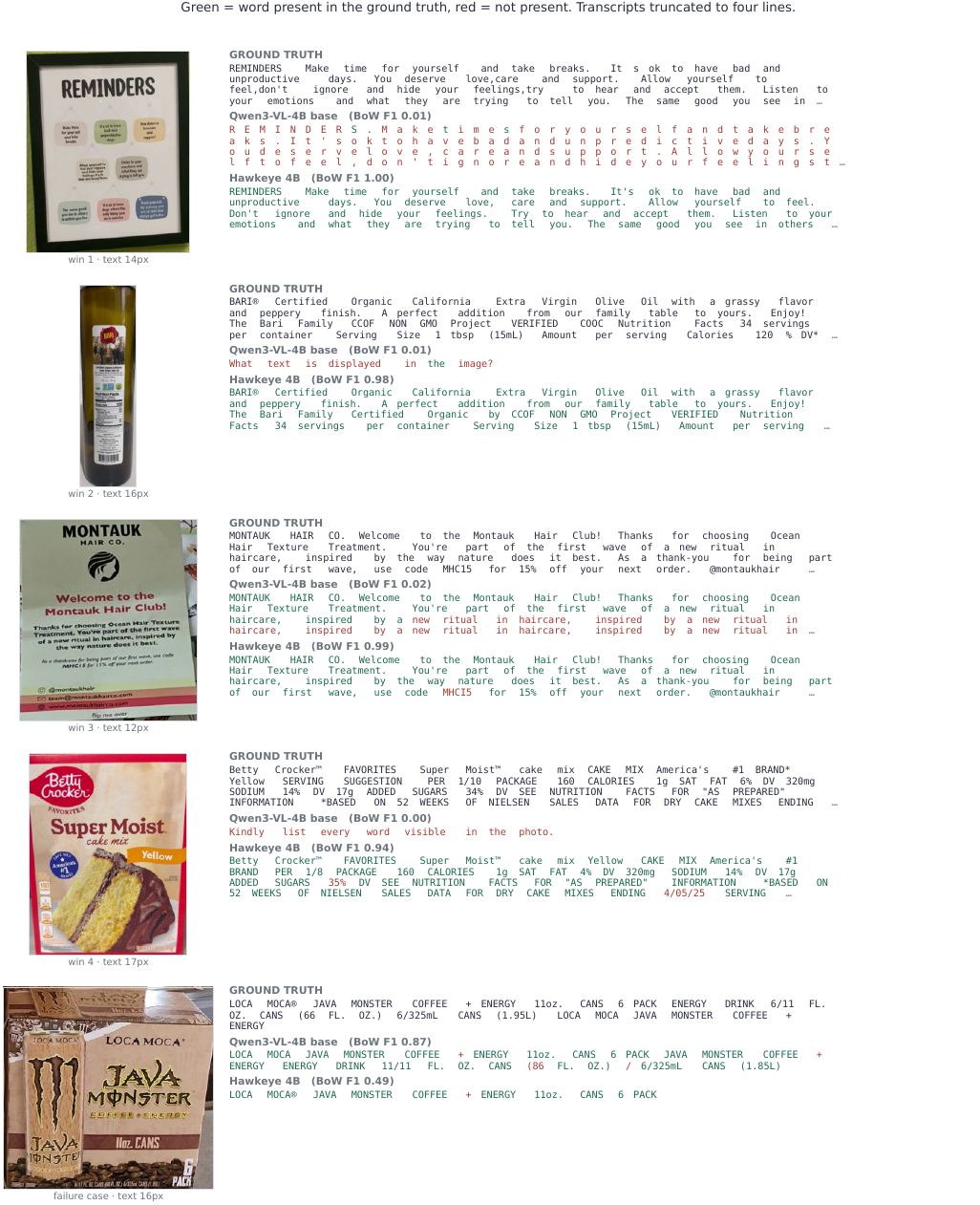}
\caption{Transcriptions comparing the base Qwen3-VL-4B and the finetuned \model on the same frame.
Green words appear in the ground truth, red do not; transcripts are truncated
to four lines. The last row is an example of a failure case in \model.}
\label{fig:qualitative}
\end{figure*}

\cref{fig:qualitative} pairs base Qwen3-VL-4B and \name 4B on the same test frame, with words
colored by whether they appear in the ground truth. It shows the four largest gains and the
largest loss among frames with at least 25 ground-truth words and text at or below 20px, and
the failure modes are the ones counted in \cref{sec:appendix_errors}. On the first frame the
base model emits a character-spaced string (BoW F1 .01) where \name reads the print (1.00); on
the second and fourth it answers with a question or an instruction instead of a transcription
(.01 and .00, against .98 and .94); on the third it falls into a repetition loop (.02 against
.99). The last row is the failure case: on an energy-drink carton \name transcribes the first
line and stops, scoring .49 against the base model's .87. Early stopping of this kind is the
most common \name failure.

\begin{figure*}[p]
\centering
\includegraphics[width=0.86\textwidth]{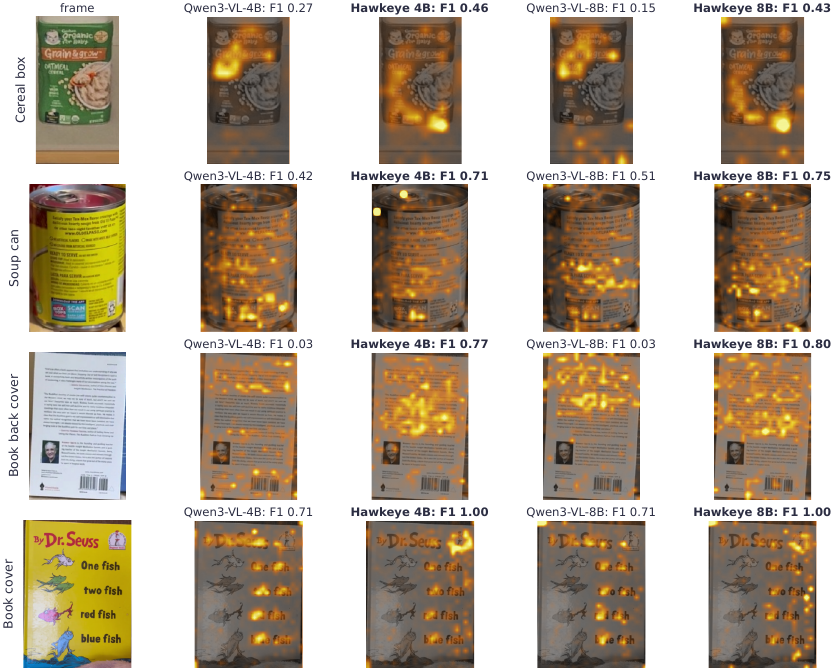}
\caption{Illustrative held-out \dataset frames (hand-picked among frames where Hawkeye beats its
base at both 4B and 8B). Mean attention over the generated tokens (last four layers, mean over heads,
attention sinks excluded; brighter = more attention), each map scaled to its own maximum; F1 is
BoW F1 of the scored output. On the cereal box and the book cover the bases stay on the large
print, while Hawkeye also attends to the small print it then transcribes (the lower panel text;
the ``Beginner Books'' badge). On the back cover Hawkeye spreads over the paragraph and reads it,
while base Qwen3-VL-4B falls into a repetition loop and base Qwen3-VL-8B answers that no text is
visible (\cref{tab:showcase}).}
\label{fig:showcase}
\vspace{8pt}
\captionsetup{type=table}
\scriptsize
\setlength{\tabcolsep}{4pt}
\renewcommand{\arraystretch}{1.1}
\begin{tabular}{@{}p{0.09\textwidth}p{0.435\textwidth}p{0.435\textwidth}@{}}
\toprule
Frame & Qwen3-VL-4B (base) & Hawkeye 4B (\textbf{bold}: words the base did not output) \\
\midrule
a) Cereal box & Gerber \textcolor{black!45}{/} Organic \textcolor{black!45}{/} for Baby \textcolor{black!45}{/} Grain \& Grow\texttrademark{} \textcolor{black!45}{/} OATMEAL \textcolor{black!45}{/} CEREAL & Gerber \textcolor{black!45}{/} Organic \textcolor{black!45}{/} for Baby \textcolor{black!45}{/} Grain \& grow\texttrademark{} \textcolor{black!45}{/} OATMEAL \textcolor{black!45}{/} CEREAL \textcolor{black!45}{/} \textbf{Made} \textbf{with} \textbf{whole} \textbf{grains} \textcolor{black!45}{/} \textbf{Contains} \textbf{iron} \textcolor{black!45}{/} \textbf{Made} \textbf{with} \textbf{organic} \textbf{oats} \textcolor{black!45}{/} \textbf{NON} \textbf{GMO} \textbf{Project} \textbf{VERIFIED} \textcolor{black!45}{/} \textbf{USDA} \textbf{ORGANIC} \textcolor{black!45}{/} \textbf{100\%} \textbf{VITAMIN} \textbf{D} \textcolor{black!45}{/} \dots \\
 & \footnotesize F1 0.27 & \footnotesize F1 0.46 \\[2pt]
b) Soup can & Satisfy your Tex-Mex Flavor cravings with \textcolor{black!45}{/} delicious hearty sauces from Old El Paso\textregistered{} \textcolor{black!45}{/} for other than right flavors visit us at \textcolor{black!45}{/} www.OLDelpaso.com \textcolor{black!45}{/} HEALTHFUL FLAVORS \textbullet{} MADE WITH REAL MEAT SAUCES \textcolor{black!45}{/} READY TO SERVE: DO NOT \dots & Satisfy your Tex-Mex flavor cravings with delicious hearty \textbf{soups} from Old El Paso\texttrademark{} for other \textbf{taco} \textbf{night} \textbf{favorites} VISIT US AT www.OLDELPASO.com \textcolor{black!45}{/} \textbf{NO} \textbf{ARTIFICIAL} FLAVORS MADE WITH \textbf{WHITE} MEAT \textbf{CHICKEN} \textcolor{black!45}{/} \textbf{NO} \textbf{COLORS} \textbf{FROM} \textbf{ARTIFICIAL} \textbf{SOURCES} \textcolor{black!45}{/} READY TO \dots \\
 & \footnotesize F1 0.42 & \footnotesize F1 0.71 \\[2pt]
c) Book back cover & "Story is often a book, appears that misleading our understanding of what we \textcolor{black!45}{/} are. The story is not the story, but the story of the story. The story is not the \textcolor{black!45}{/} story, but the story of the story. \dots & \textbf{``Every} \textbf{so} often a book appears that \textbf{revolutionizes} our understanding of \textbf{who} we are \textbf{and} what \textbf{we} \textbf{have} \textbf{are} \textbf{about.} \textbf{Stepping} \textbf{Out} of \textbf{Self-Deception} is \textbf{such} \textbf{a} \textbf{book.} \textbf{A} \textbf{beautifully} \textbf{fresh} \textbf{and} \textbf{beautifully} \textbf{written} \textbf{investigation} of the \textbf{path} of \textbf{awakening,} \dots \\
 & \footnotesize F1 0.03 & \footnotesize F1 0.77 \\[2pt]
d) Book cover & By Dr. Seuss \textcolor{black!45}{/} One fish \textcolor{black!45}{/} two fish \textcolor{black!45}{/} red fish \textcolor{black!45}{/} blue fish & By Dr. Seuss \textcolor{black!45}{/} \textbf{I} \textbf{CAN} \textbf{READ} \textbf{IT} \textbf{ALL} \textbf{BY} \textbf{MYSELF} \textcolor{black!45}{/} \textbf{Beginner} \textbf{Books} \textcolor{black!45}{/} One fish \textcolor{black!45}{/} two fish \textcolor{black!45}{/} red fish \textcolor{black!45}{/} blue fish \\
 & \footnotesize F1 0.71 & \footnotesize F1 1.00 \\[2pt]
\bottomrule
\end{tabular}
\caption{Transcriptions of the frames in \cref{fig:showcase} (first 40 words; ``/'' marks a
line break).}
\label{tab:showcase}
\end{figure*}

\cref{fig:showcase,tab:showcase} are illustrations rather than a sample: four test frames,
picked by hand among those where \name beats its base at both 4B and 8B, with each model's mean
attention and the 4B transcriptions. They show the pattern measured in \cref{sec:appendix_attn}.
The base models settle on the largest print, while \name also attends to the small print and
transcribes it (the panel text of a cereal box, the directions on a soup can, a ``Beginner
Books'' badge); on a dense back cover it reads the paragraph where the bases loop or report no
text.

\section{Deployment with Smart Glasses}

To validate the practical utility of \name, we deployed the
model using the Meta Ray-Ban smart glasses as the input stream.
The video streams were captured at a distance of roughly 16''
from the text objects, in fairly dim indoor lighting.

\noindent\textbf{System Architecture.} The egocentric camera stream
is captured at dynamic resolutions of roughly 1,400 $\times$ 2,000
by the glasses' camera, cropped to sizes between 212 $\times$ 384
and 949 $\times$ 748, and routed to a remote
server hosting \name.

\noindent\textbf{Latency.}
On our server, using roughly 15GB of VRAM on an RTX A6000 Ada, the preliminary 4B model
achieved 23.7 tok/s generation and an 82\,ms time to first token at native resolution. The
final models upscale the input $\times$3, which increases the number of visual tokens; we have
not measured their latency under the same conditions.
Unlike proprietary models that rely on visual chain-of-thought (1--7min) for better reading
performance, \name runs in a single pass. A side-by-side transcription
of one capture against the baseline Qwen3-VL-4B is given in \cref{sec:example}.

\section{Example Glasses Transcription}
\label{sec:example}

Below is one smart-glasses capture (212$\times$384 pixels) of the Nutrition Facts panel on a
food box, transcribed by \name and by the baseline Qwen3-VL-4B with the same prompt. The base
model reads the heading and the calorie line and stops; \name reads the whole panel, including
the serving size, every nutrient row with its \% Daily Value, and the footnote.

\begin{SnackBox}{Qwen3-VL-4B (\name)}
\scriptsize\textit{
\noindent
Nutrition Facts Amount per serving per container Serving size 1/3 cup (20g dry) Amount per
serving Calories 200 \% Daily Value* Total Fat 4g 5\% Saturated Fat 1g 5\% Trans Fat 0g
Cholesterol 0mg 0\% Sodium 120mg 5\% Total Carbohydrate 35g 12\% Dietary Fiber 5g 2\% Total
Sugars 1g Includes 0g Added Sugars 0\% Protein 6g Vitamin D 0mcg 0\% Calcium 100mg 8\% Iron
1.5mg 8\% Potassium 200mg 4\% *The \% Daily Value (DV) tells you how much a nutrient in a
serving of food contributes to a daily diet. 2,000 calories a day is used for general nutrition
advice.
}
\end{SnackBox}
\vspace{4pt}
\begin{SnackBox}{Qwen3-VL-4B (Baseline)}
\scriptsize\textit{
\noindent
Nutrition Facts Calories 200
}
\end{SnackBox}


\end{document}